\documentclass{article}

\usepackage[preprint]{corl_2026} % Uncomment for pre-prints (e.g., arxiv); This is like ``final'', but will remove the CORL footnote.
\usepackage{bm}
\usepackage{xcolor}
\usepackage{amsmath}
\usepackage{amssymb}
\usepackage{booktabs}
\usepackage{graphicx}
\usepackage{caption}
\usepackage{placeins}
\usepackage{cleveref}
\usepackage{siunitx}
\usepackage{comment}
\usepackage{todonotes}
\usepackage{subcaption}

\title{Embodiment-conditioned Generalist Control for Multirotor Aerial Robots}

\author{
Orestis Konstantaropoulos \hspace{0.8em} Welf Rehberg \hspace{0.8em} Mihir Kulkarni \hspace{0.8em} Kostas Alexis\\
Department of Engineering Cybernetics\\
Norwegian University of Science and Technology (NTNU), Trondheim, Norway\\
\texttt{orestiskonsta@gmail.com},
\texttt{welf.rehberg@ntnu.no}, \\
\texttt{mihir.kulkarni@ntnu.no}, \texttt{konstantinos.alexis@ntnu.no}
}

\begin{document}
\maketitle

%===============================================================================

\begin{abstract}
    We present a generalist position control policy capable of controlling arbitrary multirotor configurations of a certain rotor count (e.g., hexarotors or quadrotors) with a single set of network weights. The policy is conditioned on a physics-grounded embodiment descriptor: a mass and inertia-normalized control allocation matrix that captures how mass-normalized motor thrusts generate linear and angular accelerations in the body-frame. To train the policy, we sample from a broad distribution of arbitrary multirotor configurations, including non-planar and asymmetric systems, and optimize a single, compact network using Proximal Policy Optimization. Training requires only five minutes on an RTX $3090$ GPU using a custom NVIDIA Warp-based dynamics simulator. Through extensive simulation experiments, we show that embodiment conditioning enables robust generalist control across arbitrary morphologies. We demonstrate zero-shot real-world transfer of this generalist policy on three diverse hexarotor systems, including a planar robot, a partially symmetric non-planar system, and a random asymmetric, non-planar configuration.
\end{abstract}


\vspace{-0.5em}
\noindent\begin{center}
\includegraphics[width=\linewidth]{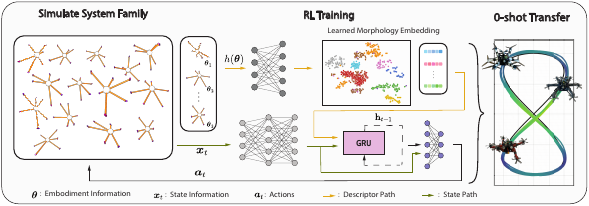}
\captionof{figure}{We train a generalist policy in simulation using a broad distribution of multirotor systems (e.g., of the hexarotor class). The policy is conditioned on both the state and an explicit model-based embodiment descriptor and directly outputs motor commands. We demonstrate real-world zero-shot transfer on substantially different configurations, including asymmetric, and non-planar designs.}
\label{fig:beauy_figure}
\end{center}
\vspace{-1em}

% \textcolor{red}{
% TODO:
% \begin{itemize}
%     \item 
% \end{itemize}
% }

%===============================================================================
% \begin{figure}[!t]
%     \centering
%     \includegraphics[width=\linewidth]{pics/beauty_figure.pdf}
%     \caption{We train a generalist policy in simulation using a broad distribution of multirotor systems of the hexarotor class. The policy is conditioned on both state and explicit model-based embodiment information and directly outputs motor commands. We demonstrate zero-shot transfer to the real world on substantially different configurations, including non-symmetric, and non-planar designs.}
%     \label{fig:beauy_figure}
% \end{figure}
% \FloatBarrier

\section{Introduction}
Multirotors are widely used due to their agility and simple design. Recent advances in onboard autonomy have enabled their deployment across a variety of environments~\cite{pretto2020building,chung2023into,dharmadhikari2025semantics}, including for drone racing with performance levels surpassing expert human pilots~\cite{kaufmann_champion-level_2023}. 
Due to their simplicity, planar multirotor configurations are dominating the field. However, research has highlighted the potential benefits offered by unconventional non-planar configurations~\cite{rajappaModelingControlDesign2015,garanger2025dodecacopter,franchi2025n,rashad2020fully,shenModelingControlOmnidirectional2022,ollero2021past}. Even within planar designs, changes to components such as motors or propellers can significantly affect control performance and often require retuning the control cascade. These challenges are amplified in non-planar and asymmetric systems, since commonly used controllers (e.g.,~\cite{lee_geometric_2010}) often rely on assumptions such as vertical thrust and partial design symmetry.

Learning-based control can, in principle, reduce the manual effort required to tune a controller for a given system. However, currently this still requires training a bespoke policy tailored to the specific robot configuration at hand. Lately, a niche set of works has started investigating generalist policies across broader families of systems. In legged locomotion, recent work has explored policies conditioned on embodiment or system parameters to enable control across different morphologies~\cite{bohlinger2025policyrunallendtoend}. For aerial robots, however, learning a single controller that generalizes across substantially different multirotor configurations remains largely unexplored. Heading in that direction, RAPTOR~\cite{raptor} recently presented a general control, which is, however, limited to planar quadrotors in a symmetric ``X'' configuration.
To overcome these limitations, we propose an embodiment-conditioned control generalist for arbitrary multirotor systems. The policy is conditioned on a compact, physics-informed descriptor of the platform morphology and is trained over a broad distribution of either quadrotor or hexarotor configurations, including planar, non-planar, symmetric, and fully asymmetric designs with vastly different scale, mass, and inertia. This enables a single generalist policy to adapt its control behavior to the robot's actuation structure and dynamics. An overview of our method is shown in \Cref{fig:beauy_figure}. 
Our analysis demonstrates that such explicit embodiment conditioning enables robust control across diverse multirotor morphologies. We train a single generalist controller in simulation in under five minutes on a single RTX 3090 GPU and demonstrate generalization across broad families of multirotors and zero-shot transfer to representative planar, non-planar, and asymmetric real-world embodiments. 
The source code for the simulator, training pipeline, and policy implementation is available at \url{https://github.com/ntnu-arl/generalist_multirotor_control.git}.

\section{Related Work}
The idea of conditioning robot control policies on embodiment information has lately received attention. Recent work in this direction conditions policies on robot morphology using graphs \cite{luo2025gcntgraphbasedtransformerpolicies}, hypernetworks \cite{xiong2023universalmorphologycontrolcontextual}, or transformers \cite{luo2025gcntgraphbasedtransformerpolicies, bohlinger2025policyrunallendtoend,sferrazza2024bodytransformerleveragingrobot}, mainly in articulated locomotion settings. These methods assume a kinematic structure composed of discrete, articulated modules, which is not a natural representation for multirotor systems where control relies on the force-torque coupling of a single rigid body. Looking at flying robots, the prevailing approach to learning for flight control involves the training of bespoke policies~\cite{eschmann_learning_2024, hwangbo_control_2017}. Toward learning multirotor flight control applicable to a relatively wide range of systems, a demonstrated approach is domain randomization over dynamic parameters and training a single policy across varying system dynamics~\cite{de_croon, molchanov_SimToMultiReal}. While effective for variations in mass or thrust coefficients and system scale, these methods assume a fixed motor layout, thus not allowing for arbitrary configurations. An alternative is adaptive control via system identification~\cite{loquercio_adaptation}, where a separate module infers system parameters from history and the policy conditions on these estimates. This enables adaptation to varying dynamics but remains focused on fixed multirotor geometries. A step toward more significant generalization, RAPTOR~\cite{raptor} instead trains separate teacher policies for varying sampled planar quadrotors in a symmetric X configuration with purely vertical thrust and distills them into a single student. Yet RAPTOR remains limited to such symmetric, planar quadrotors. In contrast, we focus on training a single morphology-conditioned policy across diverse planar and non-planar, symmetric and asymmetric multirotor morphologies. To that end, we propose the use of an explicit model-based encoding of an embodiment via a normalized allocation matrix and parameters defining the actuator dynamics. We present results both for hexarotors (owing to their rich configuration space) and quadrotors (with details in Appendix~\ref{app:quadrotors}).

% We propose the use of an explicit model-based encoding of an embodiment via a normalized allocation matrix and parameters defining the actuator dynamics, thus enabling generalization across diverse multirotor configurations.

% \textcolor{red}{
% TODO
% \begin{itemize}
%     \item 
% \end{itemize}
% }
\section{Problem Formulation}

This work addresses the problem of controlling a wide distribution of multirotor systems from the same morphology class (e.g., hexarotors or quadrotors) using a single generalist policy. Our goal is to find a \emph{single} policy able to generalize with the same parameters across vastly diverse planar and non-planar, symmetric and asymmetric multirotors, and demonstrate real-world zero-shot transfer. 

Let $\Theta$ denote a family of multirotor systems with $n_m$ motors, where each system is parameterized by a morphology vector $\bm{\theta} \in \Theta$. Importantly, beyond conventional planar systems, non-planar and asymmetric designs are considered. The morphology vector $\bm{\theta}$ contains all physical parameters required to uniquely define the platform-specific dynamics used in simulation. We formulate the control problem as a Partially Observable Markov Decision Process (POMDP) defined by the tuple $(\mathcal{S}, \mathcal{A}, p_T, \bm{s}_0, r, \gamma, \mathcal{O})$. The state space $\mathcal{S}$ contains states $\bm{s}_t \in \mathcal{S}$ comprising the system position $\bm{p}_t \in \mathbb{R}^3$ in the world frame $\mathcal{W}$, the orientation quaternion $\bm{q}_t \in \mathbb{R}^4$ with $\|\bm{q}_t\| = 1$, the linear velocity $\bm{v}_t \in \mathbb{R}^3$ expressed in $\mathcal{W}$, the angular velocity $\bm{\omega}_t \in \mathbb{R}^3$ expressed in the body frame $\mathcal{B}$, and the propeller speeds $\bm{\Omega}_t \in \mathbb{R}^{n_m}$. In addition, the morphology parameters $\bm{\theta}$ and the previous action $\bm{a}_{t-1}$ are included in the state representation, yielding the augmented state
$
\bm{s}_t =
[\bm{p}_t,\bm{q}_t,\bm{v}_t,\bm{\omega}_t,\bm{\Omega}_t,\bm{a}_{t-1},\bm{\theta}]^T.
$
The previous action is included as it is required for reward computation. Although $\bm{\theta}$ remains constant throughout an episode, we include it in the state representation to preserve the Markov property and enable conditioning of the policy on the platform morphology. The action space is denoted by $\mathcal{A}$, where the action $\bm{a}_t = \bm{\Omega}^{*}$ corresponds to the desired propeller speeds. The system dynamics are described by the transition probabilities $p_T(\bm{s}_{t+1} \mid \bm{s}_t, \bm{a}_t)$. During training, morphologies are sampled from a distribution $p_{\Theta}$ over the system family. Furthermore, $\bm{s}_0$ denotes the initial state distribution, $r(\bm{s}_t,\bm{a}_t)$ defines the reward function and $\gamma \in [0,1)$ is the discount factor. The observation space $\mathcal{O}$ contains the subset of state variables observable by the policy. Specifically, the observations $\bm{o}_t \in \mathcal{O}$ contain all state components except the past action and the motor speeds $\bm{\Omega}$, since real platforms often do not provide motor speed feedback, thereby rendering the process partially observable. A summary of the notation used in this work is provided in \Cref{tab:notation}. Given the morphology distribution $p_{\Theta}$, our objective is to learn a \emph{single} control policy that generalizes across the entire family of systems:
\vspace{-0.1cm}
\begin{equation}
    \pi^\star \in 
    \arg\max_{\pi}
    \;
    \mathbb{E}_{\bm{\theta} \sim p_{\Theta}}
    \left[
    J(\pi,\bm{\theta})
    \right], \quad \text{where} \quad
    J(\pi,\bm{\theta})
    =
    \mathbb{E}
    \left[
    \sum_{t=0}^{T}
    \gamma^t r(\bm{s}_t,\bm{a}_t)
    \right].
\end{equation}
with $\bm{a}_t \sim \pi(\bm{o}_t)$ and $\bm{s}_{t+1} \sim p_T(\cdot \mid \bm{s}_t,\bm{a}_t)$, 

% \textcolor{red}{
% TODO:
% \begin{itemize}
%     \item 
% \end{itemize}}
\FloatBarrier
\section{Methodology}\label{sec:method}

This section presents the proposed approach for generalist multirotor flight control. For the sake of derivation and presentation, the class of hexarotors is considered. 

\subsection{State Transition Dynamics}
The state transition probabilities $p_T$ of the POMDP of an arbitrary multirotor system are governed by rigid-body Newton-Euler equations and a motor model as shown in \Cref{eq:om_dot}. Note that the embodiment parameters $\bm{\theta}$ are constant during an episode. A first-order motor model is considered.
\begin{center}
  \begin{minipage}[b]{.45\textwidth}
  \begin{equation}\nonumber
      \dot{\bm{p}} = \bm{v} 
      \label{eq:p_dot}
  \end{equation}
    %\smallskip
  \begin{equation}\nonumber
      \dot{\bm{v}} = \frac{1}{m}\bm{R}(\bm{q})\bm{f}_\mathcal{B} + \bm{g}
      \label{eq:v_dot}
  \end{equation}
  %\smallskip
  \begin{equation}\nonumber
      \dot{\bm{\Omega}} = \text{diag}(\bm{\kappa})^{-1}(\bm{\Omega}^{*} - \bm{\Omega})
      \label{eq:Omega_dot}
    \end{equation}
  \end{minipage}
  \quad 
  \begin{minipage}[b]{.45\textwidth}
    \begin{equation}\nonumber
      \dot{\bm{q}} = \frac{1}{2}\bm{q} \otimes 
      \left(\begin{array}{c} 
      0 \\
      \bm{\omega}_B  \\
      \end{array}\right)
      \label{eq:q_dot}
    \end{equation}
      %\smallskip
    \begin{equation}
      \dot{\bm{\omega}}_\mathcal{B} = \bm{J}^{-1}(\bm{\tau}_\mathcal{B} - \bm{\omega} \times \bm{J} \bm{\omega})
      \label{eq:om_dot}
    \end{equation}
        %\smallskip
    \begin{equation}\nonumber
      \dot{\bm{\theta}} = 0.
      \label{eq:theta_dot}
    \end{equation}
  \end{minipage}
\end{center}
Here $m$ is the system mass, $\bm{J}$ its inertia, and $\bm{\kappa}$ the actuator time constants. The commanded rotor velocities are denoted by $\bm{\Omega}^{*}$. The rotation matrix $\bm{R}(\bm{q})\in\mathrm{SO}(3)$ maps body-frame vectors to the inertial frame,
$\mathcal{B}\!\to\!\mathcal{W}$, while $\bm{g} \approx [0, 0, -9.81]$ \si{\meter\per\square\second} denotes acceleration due to gravity. The body-frame wrench is composed of thrust
$\bm{f}_{\mathcal{B}}$ and torque $\bm{\tau}_{\mathcal{B}}$, defined through the allocation matrix $\bm{B}\in\mathbb{R}^{6\times 6}$ with $[\bm{f}_{\mathcal{B}}, \bm{\tau}_{\mathcal{B}}]^{T} = \bm{B}\bm{f}$, where $\bm{f}\in\mathbb{R}^{6}$ are the individual motor forces, with $f_i = c_{t_i} \Omega_i^2$ and $c_{t_i} \in \mathbb{R}$ being the thrust coefficient of rotor $i$. The allocation matrix is given by
\begin{equation}
    \mathbf{B}=\left[
  \begin{matrix}
    \mathbf{R}_1  \bm{z}_{1} & ... & \mathbf{R}_{6}  \bm{z}_{6}\\ 
    \bm{t}_1 \times \mathbf{R}_1 \bm{z}_{1} - \alpha_1 c_{q_i}\mathbf{R}_1 \bm{z}_{1} & ... &
    \bm{t}_{6} \times \mathbf{R}_{6} \bm{z}_{6} - \alpha_{6} c_{q_i}\mathbf{R}_{6} \bm{z}_{6}
  \end{matrix}\right].
\end{equation}
where $\bm{z}_{i}$ is the thrust direction in the motor frame $\mathcal{M}_i$ ($\bm{z}_{i}=[0,0,1]^T$ for all motors), $\bm{t}_i$ is the rotor position relative to the center-of-mass, $\mathbf{R}_i$ is the rotation matrix defining the orientation of motor $i$ with respect to the body frame, $\alpha_i\!\in\!\{-1,+1\}$ is the rotor spin direction, and $c_{q_i}$ the
torque-to-thrust ratio of rotor $i$. To support large-scale reinforcement learning, we developed a GPU-accelerated simulator using NVIDIA Warp~\cite{warp2022} that evolves $N_{\mathrm{env}}$ instances. Environment instances are simulated in parallel, each with a unique robot morphology, with batched observations across environments, and throughput exceeding $10^{9}$ steps per second on a laptop GPU. 
%As with the presented method, this simulator will be open-sourced upon acceptance of this manuscript.

% \textcolor{red}{
% TODO:
% \begin{itemize}
%     \item we don't have the evolution of $a_{t-1}$ in here
% \end{itemize}}

\subsection{Morphology Distribution}
% \FloatBarrier

\begin{figure}[t]
    \centering

    \begin{minipage}[t]{0.51\linewidth}
        \centering
        \vspace{0cm}
        \begin{minipage}[t]{0.63\linewidth}
            \centering
            \includegraphics[width=\linewidth]{ 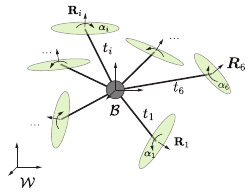}
            % \captionof{figure}{}
        \end{minipage}
        \hfill
        \begin{minipage}[t]{0.3\linewidth}
            \centering
            \includegraphics[width=\linewidth]{ 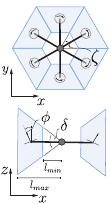}
            % \captionof{figure}{}
        \end{minipage}

        \captionof{figure}{Left: description of the system family considered in this work and limits for motor positions. Right: feasible regions are shown as light blue shaded areas.}
        \label{fig:system_overview}

    \end{minipage}
    \quad \quad
    \begin{minipage}[t]{0.41\linewidth}
        \centering
        \vspace{0.8cm}
        \captionof{table}{Parameter ranges for morphology randomization.}
        \scriptsize
        \begin{tabular}{ll}
        \toprule
        \textbf{Parameter} & \textbf{Value} \\
        \midrule
        Mass $m$ & [0.02,\:2.0]\:\si{\kilogram} \\
        Arm length $l_{\min}$ & \SI{0.1}{\meter} \\
        Arm length $l_{\max}$ & \SI{0.3}{\meter} \\
        Tilt angle $\phi$ & \SI{60}{\degree} \\
        Tilt direction $\zeta, \delta$ & \SI{60}{\degree},\;\SI{90}{\degree} \\
        Thrust coefficient $c_t$ & [$10^{-5}$,\;$5\times10^{-5}$] \\
        Motor time constant $\kappa$ & $[0.01,\; 0.06]$ \si{\second} \\
        Torque-to-thrust ratio $c_q$ & $[0.01,\; 0.05]$ \\
        \bottomrule
        \end{tabular}
        \label{tab:drone_params}
    \end{minipage}

\end{figure}

To train the generalist policy, we sample morphologies (shown in \Cref{fig:system_overview}) from a distribution spanning variations in geometry, mass and inertial properties, actuator dynamics, and thrust characteristics. Each sample is fully described by its parameter vector
\begin{gather}
    \theta = [m, \textrm{vec}(\bm{J}), \bm{t}_1,...,\bm{t}_{6}, \textrm{vec}(\bm{R}_1),..., \textrm{vec}(\bm{R}_{6}), \bm{c}_{t}, \bm{c}_{q}, \bm{\kappa}].
\end{gather}
where $\textrm{vec}()$ represents the operator to vectorize the elements of a matrix. Not all randomly generated morphologies correspond to physically meaningful or controllable systems. To limit the distribution to meaningful multirotor configurations we sample systems within the limits described in \Cref{tab:drone_params} and \Cref{fig:system_overview}. Note that the motor thrust direction is parameterized only by $\phi$, the angle with respect to the $z$-axis of $\mathcal{B}$, since rotations about the $z$-axis (yaw) are unrestricted. Additionally, we apply morphology feasibility filtering before training. First, sampled platforms must satisfy a hover feasibility condition: being able to produce total thrust higher than the gravitational force while keeping zero moment, i.e. $||\bm{f}^{\max}_{\mathcal{B}}||_2 \ge mg, \bm{\tau}_{\mathcal{B}}=0$. Second, the maximum achievable thrust along the body-frame thrust axis is required to exceed the vehicle weight by a factor $b_f$, i.e., $|f^{\max}_{\mathcal{B},z}| \geq b_f mg$, ensuring that the platform can sustain hover with margin. Third, we enforce a minimum control authority skill by requiring a minimum angular acceleration around each axis.

% \textcolor{red}{
% TODO:
% \begin{itemize}
%     \item 
% \end{itemize}}

\subsection{Embodiment Conditioning}

To enable generalization to a wide distribution of systems, the generalist policy is conditioned on embodiment information and receives morphology-dependent observations both during training and inference. A key design choice is the representation of the morphology parameters provided to the policy. A straightforward approach is to condition the network on the full parameter vector $\bm{\theta}$. While this representation is sufficient to fully specify the system dynamics, our ablation studies showed that learning improves significantly when using a lower-dimensional, control-oriented embedding $\bm{z}_{\mathrm{morph}}=h(\bm{\theta})$ that captures the control-relevant structure of the platform. Rather than directly using raw physical parameters, we represent each platform using a mass and inertia normalized allocation matrix $\bm{B}_{norm}$, derived below. Starting from the multirotor dynamics, the translational and rotational accelerations induced by the rotor forces can be written as
\vspace{-0.cm}
\begin{equation}
\dot{\bm{v}}
=
\frac{1}{m}\bm{R}(\bm{q})\bm{B}_f\bm{f}
+
\bm{g},
\end{equation}
\vspace{-0.3cm}
\begin{equation}
\dot{\bm{\omega}}
=
\bm{J}^{-1}\bm{B}_\tau\bm{f}
-
\bm{J}^{-1}
(
\bm{\omega}\times\bm{J}\bm{\omega}
),
\end{equation}
where $\bm{B}_f$ and $\bm{B}_\tau$ denote the force and torque components of the allocation matrix $\bm{B}=[\bm{B}_f, \bm{B_{\tau}}]^T$. We can now write:
\begin{equation}
\begin{bmatrix}
\dot{\bm{v}} \\
\dot{\bm{\omega}}
\end{bmatrix}
=
\underbrace{
\begin{bmatrix}
\bm{R}(\bm{q}) & \bm{0} \\
\bm{0} & \bm{I}
\end{bmatrix}
}_{\text{frame mapping}}
\underbrace{
\begin{bmatrix}
\bm{B}_f \\
m\bm{J}^{-1}\bm{B}_{\tau}
\end{bmatrix}
}_{\bm{B}_{\mathrm{norm}}}
\underbrace{
\frac{\bm{f}}{m}
}_{\bm{f}_{\mathrm{norm}}}
+
\begin{bmatrix}
\bm{g} \\
-\bm{J}^{-1}
(
\bm{\omega}\times\bm{J}\bm{\omega}
)
\end{bmatrix}.
\label{eq:dynamics_full}
\end{equation}
This motivates the definition of $\bm{f}_{\text{norm}} = \bm{f}/m$ and the normalized allocation matrix as $\bm{B}_{\mathrm{norm}} = [\bm{B}_f, m\bm{J}^{-1}\bm{B}_\tau]^{T}$, mapping the normalized rotor thrusts to linear and angular accelerations to directly encode the control authority induced by motor placement, mass distribution, and torque-to-thrust ratios in the acceleration space. Finally, the morphology descriptor provided to the policy is constructed as $\bm{z}_{\mathrm{morph}} = h(\theta) =
[\textrm{vec}(\bm{B}_{\mathrm{norm}}), \bm{\kappa}, \bm{c}_t]$. This normalized descriptor enables training across heterogeneous multirotor embodiments with varying geometry and dynamics.

% \textcolor{red}{
% TODO:
% \begin{itemize}
%      \item perhaps show the difference between the embodiment representations in a pca/t-sne
%      plot
% \end{itemize}}

\subsection{Policy Architecture}
 The generalist policy receives both state and morphology information and process them separately by a state and embodiment encoder. The state encoder consists of two fully connected layers with $50$ nodes each, while the embodiment encoder consists of one fully connected layer with $50$ nodes. All hidden layers use ELU activations. The resulting latent states are concatenated and fed into a gated recurrent unit (GRU) with hidden size $50$. The GRU output is then used by three linear heads that predict the action mean, the log-standard deviation, and the value estimate. In addition, we include a skip connection from the state latent representation directly to the policy heads, allowing the policy to retain direct access to the current state alongside the recurrent representation. The policy outputs mass-normalized individual motor thrust commands, $\bm{f}_{\mathrm{norm}}^*$. These are scaled by the system mass and then converted to desired motor velocities. Then they are applied as desired motor velocities $\bm{a}_t =\bm{\Omega}^{\star}$ in the simulator. The described architecture is shown in Appendix~\ref{app:architecture}. 

% \textcolor{red}{
% TODO:
% \begin{itemize}
%     \item 
% \end{itemize}}

\subsection{Training Procedure}
\label{sec:training_procedure}

The policy is trained using Proximal Policy Optimization (PPO) with samples from the whole morphology distribution. We simulate $N_{\mathrm{env}}=2048$ environments in parallel, each corresponding to a distinct embodiment, and collect rollouts of length $H=32$ with a simulation time step $\Delta t = 0.01$\si{\second}. Episodes terminate after a maximum horizon of $600$ steps. Since different morphologies exhibit varying dynamical characteristics and stabilization behavior, we apply an adaptive sampling strategy during training to balance their contribution to the training data distribution. Morphologies with lower recent returns are sampled more frequently, increasing the amount of training data collected from configurations that are more challenging for the current policy and thereby improving performance consistency across the morphology distribution.

\vspace{-0.2cm}
\textbf{Reward Design} We define a smooth reward composed of errors for position, heading,  velocity, angular velocity, angular velocity increment, and action regularization terms. Let
\begin{gather}
e_{\bm{p}} = \|\bm{p}_t - \bm{p}^\star\|, \;\;
e_{\psi} = \psi_t - \psi^\star, \;\;
e_{\bm{v}} = \|\bm{v}_t-\bm{v}^\star\|, \;\;
e_{\bm{\omega}} = \|\bm{\omega}_t\|, \;\;
e_{\Delta \bm{\omega}} = \|\bm{\omega}_t - \bm{\omega}_{t-1}\|
\end{gather}
where $\bm{p}^\star$, $\psi^\star$ and $\bm{v}^\star$ being the desired position, heading (yaw) and velocity.  Using the defined errors we employ Gaussian shaping functions to calculate the individual reward terms
\begin{gather}
\rho(x,\lambda_1,\lambda_2) = \lambda_1 \exp(-\lambda_2 x^2), \quad r_{\square} = \rho(e_{\square},\lambda_{1,{\square}}, \lambda_{2,{\square}}),
\end{gather}
with $\square$ being a placeholder for the above mentioned individual quantities and $\lambda_{1,{\square}}$ and $\lambda_{2,{\square}}$ being weights on the individual reward terms. To encourage smooth control inputs, we add an action regularization term
\vspace{-0.1cm}
\begin{gather}
r_{\Delta \bm{a}}
=
-\lambda_{\bm{a}} \|\bm{a}_t - \bm{a}_{t-1}\|_2
\end{gather}
where $\lambda_{\bm{a}}$ are the weight scaling the action difference reward term. Additionally, we define a proximity weighting term $w = \exp(-\lambda_w e_{\bm{p}})$, which increases the importance of velocity and angular stabilization close to the hover target. Here $\lambda_w$ describes a tunable weight in the exponential. Episodes are terminated when safety constraints are violated. Specifically, a crash condition is triggered if the position error, velocity error or the angular velocity become to high. Upon termination, a fixed penalty $p_{crash}$ is applied. The total reward is then defined as
\vspace{-0.cm}
\begin{gather}
    r_t =
\begin{cases}
p_{crash}, & \text{if crashed}  \\
r_{\bm{p}} + r_{\bm{v}} + r_{\bm{\omega}} + r_{\Delta \bm{\omega}} + r_{\Delta \bm{a}} +
w \left(4 r_{\bm{v}} + 16 r_{\bm{\omega}} + r_{\psi}\right),  & \text{otherwise}.
\end{cases}
\end{gather}
%\vspace{-0.05cm}
\textbf{Sim-to-Real Training Setup} 
To facilitate zero-shot sim-to-real transfer, we augment the training of the generalist deployed on real systems with observation noise and delayed state estimates. 

\section{Results and Evaluations}
We evaluate the proposed embodiment-conditioned generalist controller through a set of simulation and real-world experiments. Notably, all policies were trained in only $5$ min each on a workstation with an RTX $3090$ GPU. Implementation details are provided in Appendix~\ref{app:implementation_details}. Throughout the experimental section, we use three representative embodiments for zero-shot evaluation: a planar configuration, a non-planar partially symmetric configuration, and a fully arbitrary configuration. This section focuses on hexarotors, while key results for quadrotors are also presented. A more detailed analysis is provided in Appendix~\ref{app:quadrotors}.

\subsection{Simulation Studies}
We first evaluate generalization across morphologies in simulation. We compare three policy classes. First, \emph{bespoke} policies are trained independently for single fixed morphologies $\bm{\theta} \in \Theta$, using observations
$\bm{o}^{BS}_t = [\bm{p}_t,\bm{q}_t,\bm{v}_t,\bm{\omega}_t, \bm{\Omega}_t]$ and an MLP architecture. Note that the choice of a purely MLP based architecture for the bespoke policies is justified by numerous works like \cite{eschmann_learning_2024, hwangbo_control_2017, rehberg2026efficientknowledgetransferjumpstarting}. 
Second, \emph{uninformed generalist} policies use a recurrent architecture and are trained over the same morphology distribution as our method, but receive no explicit embodiment information, thus
$\bm{o}^{DR}_t = [\bm{p}_t,\bm{q}_t,\bm{v}_t,\bm{\omega}_t]$.
Finally, our generalist policy is trained over the same morphology distribution, but is conditioned on the proposed embodiment descriptor. The informed general policy and uninformed general policy maximize expected return over the morphology distribution $p_{\Theta}$, whereas the bespoke policies optimize performance only for a single fixed embodiment. All policies are trained with PPO and the same reward function, and hyperparameters. We evaluate the resulting policies over $N = 1024$ sampled morphologies that include arbitrary, planar and non-planar, symmetric and asymmetric hexarotors. For each embodiment $\bm{\theta}_i$, we run $K=10$ trials. At the beginning of each trial, the robot is initialized uniformly at random inside a box of side length $0.5$\si{\meter} around the initial reference position, and is then commanded to track a Lissajous trajectory with period $T_{\mathrm{traj}}=7.5$\si{\second} by providing position and velocity setpoints. We report success rate (SR), defined as the percentage of evaluated embodiments for which the policy completes the trajectory-tracking task without crashing (crash conditions are detailed in the Appendix~\ref{app:implementation_details}), together with the root mean squared errors (RMSE) of position and velocity in \Cref{tab:sim_results_compare}.
We compute one RMSE per embodiment by averaging over all trials and timesteps. We report the mean and standard deviation of RMSE$^{(i)}$ across the evaluated embodiments. The same procedure is repeated for a quadrotor distribution, and the corresponding RMSE values are additionally reported in \Cref{tab:sim_results_compare}.

\begin{figure}[htbp]
    \centering
    \begin{subfigure}{0.4\linewidth}
        \centering
        \includegraphics[width=\linewidth]{ 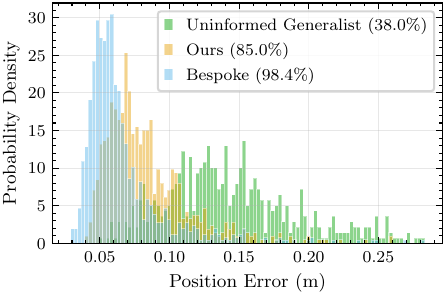}
        \caption{}
        \label{fig:rmse}
    \end{subfigure}
    \hfill
    \begin{subfigure}{0.52\linewidth}
        \centering
        \includegraphics[width=\linewidth]{ 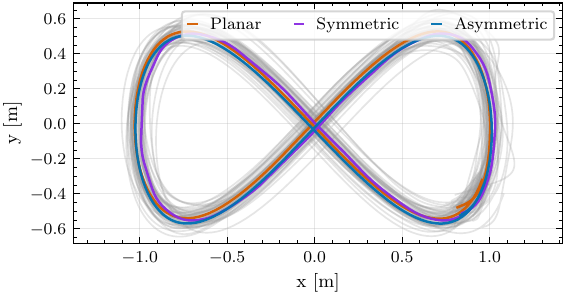}
        \caption{}
        \label{fig:traj}
    \end{subfigure}

    \caption{a) Distributions of position errors following the Lissajous trajectory for specialists, the uninformed generalist and policies trained with our method. b) Simulated trajectories of $50$ systems controlled by our generalist following a Lissajous trajectory. The systems realized for real-world tests are highlighted. Note that the simulation assumes no noise in the sensor readings or delayed measurements.}
    \label{fig:lissajous_results}
\end{figure}

\vspace{-0.2cm}
\textbf{Generalization across Morphologies} As expected, bespoke policies achieve the highest success rate, since each policy is specialized to a single fixed morphology. However, our generalist policy achieves remarkable generalization and substantially outperforms the uninformed generalist, improving the success rate from $38$\si{\percent} to $85$\si{\percent} for hexarotors and from $58$\si{\percent} to $80$\si{\percent} for quadrotors, while approaching the tracking accuracy of bespoke policies. This demonstrates that recurrence alone is insufficient to infer the relevant embodiment properties over a broad morphology distribution. Explicitly conditioning the policy on a model-based, physics-informed embodiment descriptor is critical for robust generalization. We further examine the failure cases in Appendix~\ref{app:failing_embodiments}. There, we report results on a restricted distribution of planar, non-symmetric platforms with upright motors, where the generalist achieves $100$\si{\percent} success rate and also show that failures are correlated with higher tracking errors of the corresponding bespoke policies.
In \Cref{fig:rmse}, the histograms of the position error are shown. Note that the histograms include only configurations capable of successfully tracking the trajectory. In \Cref{fig:traj}, we further visualize trajectory-tracking performance on $50$ randomly sampled morphologies, as well as zero-shot performance on the three representative embodiments. Similar results for quadrotors are detailed in the Appendix~\ref{app:quadrotors}.

\begin{table}[t]
\centering
\caption{Comparison of control policies for tracking a Lissajous trajectory showing success rate (SR) and position and velocity RMSE}
\label{tab:sim_results_compare}
\footnotesize
\setlength{\tabcolsep}{3pt}
\renewcommand{\arraystretch}{0.9}

\begin{tabular}{l|c|c|c|c|c|c}
\toprule

& \multicolumn{2}{c|}{\textbf{SR (\si{\percent})}} 
& \multicolumn{2}{c|}{\textbf{$\bm{p}$ RMSE (\si{\meter})}} 
& \multicolumn{2}{c}{\textbf{$\bm{v}$ RMSE (\si{\meter\per\second})}} \\

\midrule

\textbf{Method}
& \textbf{Hex} & \textbf{Quad}
& \textbf{Hex} & \textbf{Quad}
& \textbf{Hex} & \textbf{Quad} \\

\midrule

Bespoke Policies
& $98$ & $97$
& $0.058 \pm 0.073$ & $0.086 \pm 0.035$
& $0.053 \pm 0.114$ & $0.068 \pm 0.077$ \\

Uninformed Generalist
& $38$ & $58$
& $0.142 \pm 0.067$ & $0.181 \pm 0.080$
& $0.100 \pm 0.127$ & $0.066 \pm 0.092$  \\

\textbf{Ours}
& $85$ & $80$
& $0.082 \pm 0.037$ & $0.105 \pm 0.055$
& $0.037 \pm 0.012$ & $0.039 \pm 0.014$ \\

\bottomrule
\end{tabular}
\end{table}
\normalsize

\textbf{Ablation Studies on the Descriptor}
We next ablate the components of the proposed embodiment descriptor. We compare the full descriptor against four variants: (i) removing the thrust coefficients and the time constants (altogether called actuator constants), (ii) removing the mass--inertia normalization and providing mass and inertia as separate inputs, (iii) replacing the normalized allocation matrix with raw motor positions and orientations, and (iv) removing the embodiment descriptor entirely, corresponding to the uninformed baseline. We evaluate all variants using the same setup as above and report success rate, position RMSE, and velocity RMSE in \Cref{tab:sim_results_components}. The full descriptor achieves the highest success rate, while maintaining a low tracking error. Removing the actuator constants preserves the same success rate but increases the tracking error slightly, suggesting that the recurrent policy can partially infer actuator properties from interaction history; however, actuator constants remain important for step-response behavior and should therefore be used if available. In contrast, removing the mass-inertia normalization reduces success rate from $85$\si{\percent} to $67$\si{\percent}, and replacing the normalized allocation matrix with raw motor poses further reduces it to $54$\si{\percent}. Overall, the proposed complete physics-informed descriptor provides a more effective representation than raw dynamics and only geometric morphology information.

\begin{table}[t]
\centering
\footnotesize
\caption{Results of the ablation study for our proposed embodiment descriptor showing success rate (SR), as well as position and velocity RMSE.}
\label{tab:sim_results_components}
\renewcommand{\arraystretch}{0.9}
\setlength{\tabcolsep}{5pt}
\begin{tabular}{lcccc}
\toprule
\textbf{Method} &
\textbf{SR (\si{\percent})} &
\textbf{$\bm{p}$ RMSE (\si{m})} &
\textbf{$\bm{v}$ RMSE (\si{\meter\per\second})} \\
\midrule
full descriptor                 & 85 & $0.082 \pm 0.037$ & $0.037 \pm 0.012$\\
(i) w/o actuator constants   & 85 & $0.084 \pm 0.046$ & $0.045 \pm 0.086$\\
(ii) w/o mass--inertia normalization     & 67 & $0.081 \pm 0.045$ & $0.043 \pm 0.076$\\
(iii) raw motor poses & 54 & $0.126 \pm 0.051$ & $0.047 \pm 0.055$\\
(iv) w/o embodiment descriptor   & 38 & $0.142 \pm 0.067$ & $0.100 \pm 0.127$\\
\bottomrule
\end{tabular}
\end{table}
\normalsize

\subsection{Real-World Validation}
\begin{figure}
    \centering
    \includegraphics[width=0.61\linewidth]{ 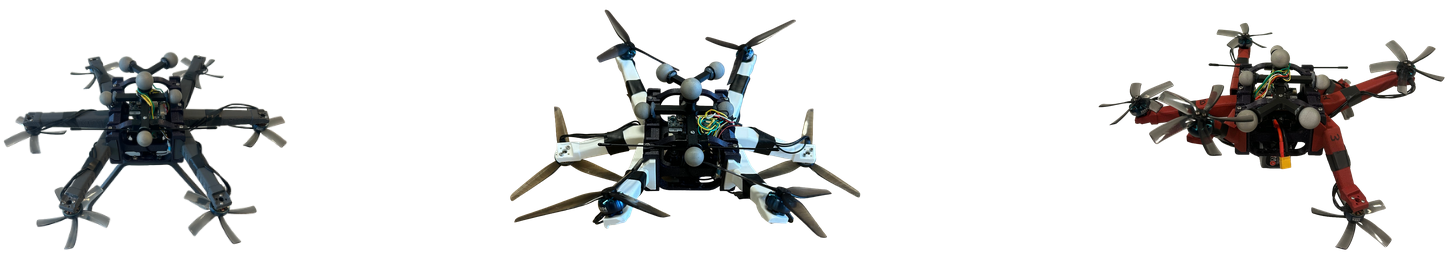}
    % \caption{Real systems used}
    \label{fig:real_systems}
    \includegraphics[width=0.86\linewidth]{ 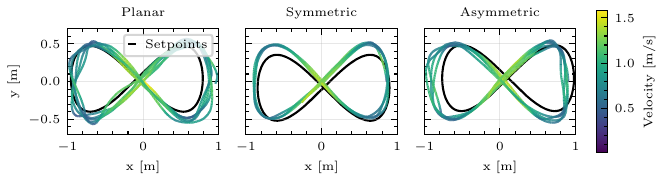}
    \caption{Top: systems used for real-world evaluation. Bottom: resulting trajectories for the real-world experiments.}
    \label{fig:real_world_experiments}
\end{figure}

To evaluate the real-world performance of our method, we deploy the informed generalist, trained incorporating the sim-to-real procedure described in \Cref{sec:training_procedure}, on three distinct systems with varying geometries, thrust coefficients, and motor time constants arising from different motor–propeller combinations, see Appendix~\ref{app:representatives}. The three platforms exhibit distinct geometries: a) a planar configuration, b) a configuration symmetric with respect to the $x$-$z$-plane and c) a random non-planar and asymmetric configuration. The systems are deployed in a Qualysis motion capture arena, providing ground truth position feedback. Each platform is equipped with a ModalAI Voxl 2 Mini autopilot board and a ModalAI Voxl ESC. The trained networks are deployed directly on the onboard compute hardware through a custom PX4 \cite{px4_autopilot} module. The systems used for the real-world evaluation are shown in \Cref{fig:real_world_experiments}. We assess the proposed method on a Lissajous trajectory tracking task with a loop period of $7.5$\si{\second}, and the resulting trajectories are presented in \Cref{fig:real_world_experiments}. The deployed generalist achieves a mean position tracking error of $0.13$\si{\meter} for the planar configuration, $0.15$\si{\meter} for the symmetric non-planar configuration and $0.21$\si{\meter} for the random non-planar and asymmetric configuration.

\section{Limitations}
While our method exhibits strong generalization within a single system family (e.g., hexarotors), it is currently designed for fixed-motor-count systems, and extending it to handle varying numbers of motors represents a promising direction for future work. Although our method achieves a high success rate ($85$\si{\percent} on hexarotors and $80$\si{\percent} on quadrotors), it still does not fully reach the near-$100$\si{\percent} reliability of bespoke policies. Further narrowing this gap by refining the adaptive morphology sampling strategy and exploring more expressive architectures capable of capturing extreme, long-tail configurations during training represents a promising research direction. Last, like in RAPTOR we aim to investigate extension to airframes with soft components. 

\section{Conclusions}
In this work, we introduced an embodiment-informed control generalist for multirotor systems. By conditioning the controller on a model-based descriptor of embodiment-relevant information, our approach achieves robust performance across a broad range of configurations and enables zero-shot transfer to real-world systems on three distinct hexarotor platforms that vary substantially in motor placement, motor orientation, inertia, and propeller type. 

% Future work will aim to extending the method to handle morphologies with varying numbers of propellers, as well as shape morphing capabilities. 
%===============================================================================

% The acknowledgments are automatically included only in the final and preprint versions of the paper.
\acknowledgments{This work was supported by the Horizon Europe Grant Agreement No. 101119774 and the Research Council of Norway under Award NO-357451. The authors are with the Department of Engineering Cybernetics, Norwegian University of Science and Technology (NTNU), Norway.}
%\clearpage
%===============================================================================

\bibliography{bibliography}  % .bib
\clearpage
\thispagestyle{empty}
\null
\appendix
%\appendix

\setcounter{figure}{0}
\setcounter{table}{0}

\renewcommand{\thefigure}{A.\arabic{figure}}
\renewcommand{\thetable}{A.\arabic{table}}

\begin{table*}[t]
    \centering
    \caption{Summary of notation used throughout the paper.}
    \label{tab:notation}
    \small
    \setlength{\tabcolsep}{5pt}
    \begin{tabular}{l|l|l|l}
        \toprule
        Symbol & Description &
        Symbol & Description \\
        \midrule

        $\Theta$ & Family of multirotor systems &
        $\bm{f}$ & Rotor thrust vector \\

        $\bm{\theta}$ & Morphology parameter vector &
        $\bm{f}_{\mathcal{B}}$ & Body-frame thrust \\

        $p_{\Theta}$ & Distribution over morphologies &
        $\bm{\tau}_{\mathcal{B}}$ & Body-frame torque \\

        $\mathcal{S}$ & State space &
        $\bm{B}$ & Allocation matrix \\

        $\mathcal{A}$ & Action space &
        $\bm{B}_f$ & Force allocation matrix \\

        $\mathcal{O}$ & Observation space &
        $\bm{B}_{\tau}$ & Torque allocation matrix \\

        $\bm{s}_t$ & Environment state at time $t$ &
        $\bm{B}_{\mathrm{norm}}$ & Normalized allocation matrix \\

        $\bm{o}_t$ & Observation at time $t$ &
        $\bm{z}_{\mathrm{morph}}$ & Morphology descriptor \\

        $\bm{a}_t$ & Action applied at time $t$ &
        $m$ & System mass \\

        $p_T$ & Transition probability distribution &
        $\bm{J}$ & Inertia matrix \\

        $r(\bm{s}_t,\bm{a}_t)$ & Reward function &
        $\bm{\kappa}$ & Motor time constants \\

        $\gamma$ & Discount factor &
        $c_t$ & Thrust coefficient \\

        $\pi$ & Control policy &
        $c_q$ & Torque-to-thrust ratio \\

        $J(\pi,\bm{\theta})$ & Expected return &
        $\bm{R}(\bm{q})$ & Body-to-world rotation matrix \\

        $T$ & Return horizon &
        $N$ & Number of sampled configurations for evaluation \\
        
        $\bm{p}_t$ & Position in frame $\mathcal{W}$ &
        $\bm{R}_i$ & Orientation of motor $i$ \\

        $\bm{q}_t$ & Orientation quaternion &
        $\bm{t}_i$ & Position of motor $i$ \\

        $\bm{v}_t$ & Linear velocity &
        $\alpha_i$ & Rotor spin direction \\

        $\bm{\omega}_t$ & Angular velocity &
        $\mathcal{W}$ & World frame \\

        $\bm{\Omega}_t$ & Rotor angular velocities &
        $\mathcal{B}$ & Body frame \\

        $\bm{\Omega}^{*}$ & Desired rotor velocities &
        $\mathcal{M}$ & Motor frame \\

        $N_{\mathrm{env}}$ & Parallel environments &
        $H$ & Rollout horizon \\

        $T_{\mathrm{traj}}$ & Trajectory period &
        $p_{crash}$ & Crash penalty \\

        $b_f$ & Hover margin factor & 
        $\bm{p}^\star$ & Desired position \\

        $\psi^\star$ & Desired heading & 
        $\bm{v}^\star$ & Desired velocity \\

        $\lambda_{1,\square}, \lambda_{2,\square}$ & Gaussian shaping weights &
        $\lambda_a$ & Action regularization weight \\

        $\lambda_w$ & Proximity term weight &
        $K$ & Evaluation trials per embodiment \\

        $\bm{o}_{BSt}$ & Bespoke policy observation &
        $\bm{o}_{DRt}$ & Uninformed generalist observation \\

        $\bm{f}_{\mathrm{norm}}$ & Mass-normalized thrust &
        $\bm{f}^\star_{\mathrm{norm}}$ & Normalized thrust command \\

        \bottomrule
    \end{tabular}
\end{table*}
\section{Morphology Descriptor}

The proposed morphology descriptor $\bm{z}_{\mathrm{morph}}$ is detailed below for completeness

\begin{align}
\bm{z}_{\mathrm{morph}} &=
\left[
\mathrm{vec}(\bm{B}_{\mathrm{norm}}), \bm{\kappa}, \bm{c}_t
\right]
\\[0.8em]
\bm{B}_{\mathrm{norm}} &=\begin{bmatrix}
\bm{B}_f \\
m\bm{J}^{-1}\bm{B}_{\tau}
\end{bmatrix} \\
 &=
\Biggl[
\begin{array}{c}
\begin{bmatrix}
\mathbf{R}_1 \bm{z}_{1} & \cdots & \mathbf{R}_{6} \bm{z}_{6}
\end{bmatrix}
\\[0.6em]
m\mathbf{J}^{-1}\begin{bmatrix}
\bm{t}_1 \times \mathbf{R}_1 \bm{z}_{1}
- \alpha_1 c_{q_1}\mathbf{R}_1 \bm{z}_{1}
&
\cdots
&
\bm{t}_{6} \times \mathbf{R}_{6} \bm{z}_{6}
- \alpha_{6} c_{q_6}\mathbf{R}_{6} \bm{z}_{6}
\end{bmatrix}
\end{array}
\Biggr]
\end{align}

\begin{comment}

\begin{align}
    \bm{z}_{\mathrm{morph}} &= 
[\textrm{vec}(\bm{B}_{\mathrm{norm}}), \bm{\kappa}, \bm{c}_t]
    &= \left[
\begin{array}{c}
\left[
\begin{array}{ccc}
\mathbf{R}_1 \bm{z}_{1} & \cdots & \mathbf{R}_{6} \bm{z}_{6}
\end{array}
\right]
\\[0.5em]
\left[
\begin{array}{ccc}
\bm{t}_1 \times \mathbf{R}_1 \bm{z}_{1}
- \alpha_1 c_{q_i}\mathbf{R}_1 \bm{z}_{1}
&
\cdots
&
\bm{t}_{6} \times \mathbf{R}_{6} \bm{z}_{6}
- \alpha_{6} c_{q_i}\mathbf{R}_{6} \bm{z}_{6}
\end{array}
\right]
\end{array}
\right]
\end{align}

\begin{equation}
    \mathbf{B}=\left[
  \begin{matrix}
    \mathbf{R}_1  \bm{z}_{1} & ... & \mathbf{R}_{6}  \bm{z}_{6}\\ 
    \bm{t}_1 \times \mathbf{R}_1 \bm{z}_{1} - \alpha_1 c_{q_i}\mathbf{R}_1 \bm{z}_{1} & ... &
    \bm{t}_{6} \times \mathbf{R}_{6} \bm{z}_{6} - \alpha_{6} c_{q_i}\mathbf{R}_{6} \bm{z}_{6}
  \end{matrix}\right].
\end{equation}

\begin{bmatrix}
\bm{B}_f \\
m\bm{J}^{-1}\bm{B}_{\tau}
\end{bmatrix}

$\bm{z}_{\mathrm{morph}} = 
[\textrm{vec}(\bm{B}_{\mathrm{norm}}), \bm{\kappa}, \bm{c}_t]$
\end{comment}

\section{Implementation Details} \label{app:implementation_details}

This section provides further implementation details. 

\subsection{Reward Design}
\label{app:reward_details}

The main text describes the reward in its general form. Here we provide the numerical instantiations used in our experiments.

\paragraph{Simulation Reward}
For the simulation results reported in the main paper, we use the following reward instantiation:
\begin{align}
r_{\bm{p}} &=
\rho(e_{\bm{p}}, 3.0, 5.0)
+
\rho(e_{\bm{p}}, 6.0, 12.0), \\
r_{\bm{v}} &=
\rho(e_{\bm{v}}, 2.0, 10.0), \\
r_{\bm{\omega}} &=
\rho(e_{\bm{\omega}}, 1.0, 0.5), \\
r_{\Delta \bm{\omega}} &=
\rho(e_{\Delta \bm{\omega}}, 1.0, 0.5), \\
r_{\psi} &=
\rho(e_{\psi}, 0.75, 6.0),
\end{align}
with proximity weight
\begin{equation}
w = \exp(-5 e_{\bm{p}}),
\end{equation}
and action smoothness penalty
\begin{equation}
r_{\Delta \bm{a}} =
-0.1 \|\bm{a}_t - \bm{a}_{t-1}\|_2 .
\end{equation}
The resulting reward is
\begin{equation}
r_t =
r_{\bm{p}}
+ r_{\bm{v}}
+ r_{\bm{\omega}}
+ r_{\Delta \bm{\omega}}
+ r_{\Delta \bm{a}}
+ w \left(4 r_{\bm{v}} + 16 r_{\bm{\omega}} + r_{\psi}\right).
\end{equation}

\paragraph{Sim-to-Real Reward}
For policies transferred to the real platform, we use a slightly smoother reward. In particular, we relax the position, velocity, and angular-velocity shaping terms, while increasing the action-difference penalty
\begin{align}
r_{\bm{p}}^{\mathrm{real}} &=
\rho(e_{\bm{p}}, 3.0, 3.0)
+
\rho(e_{\bm{p}}, 4.0, 6.0), \\
r_{\bm{v}}^{\mathrm{real}} &=
\rho(e_{\bm{v}}, 2.0, 5.0), \\
r_{\bm{\omega}}^{\mathrm{real}} &=
\rho(e_{\bm{\omega}}, 1.0, 0.3), \\
r_{\Delta \bm{\omega}}^{\mathrm{real}} &=
\rho(e_{\Delta \bm{\omega}}, 1.0, 0.5), \\
r_{\psi}^{\mathrm{real}} &=
\rho(e_{\psi}, 0.75, 6.0),
\end{align}
and
\begin{equation}
r_{\Delta \bm{a}}^{\mathrm{real}}
=
-0.5 \|\bm{a}_t - \bm{a}_{t-1}\|_2 .
\end{equation}
The final reward is
\begin{equation}
r_t^{\mathrm{real}} =
r_{\bm{p}}^{\mathrm{real}}
+ r_{\bm{v}}^{\mathrm{real}}
+ r_{\bm{\omega}}^{\mathrm{real}}
+ r_{\Delta \bm{\omega}}^{\mathrm{real}}
+ r_{\Delta \bm{a}}^{\mathrm{real}}
+
\frac{1}{7} r_{\bm{p}}^{\mathrm{real}}
\left(
r_{\bm{v}}^{\mathrm{real}}
+ 8 r_{\bm{\omega}}^{\mathrm{real}}
+ r_{\psi}^{\mathrm{real}}
\right).
\end{equation}
This modification reduces the incentive to aggressively minimize small tracking errors in simulation and instead biases the policy toward smoother control actions, which we found important for robust real-world deployment.

\paragraph{Termination}
For both reward instantiations, episodes terminate when
\begin{equation}
e_{\bm{p}} > 4.0\text{\si{\meter}}, \qquad
e_{\bm{v}} > 20.0\text{\si{\meter\per\second}}, \qquad
e_{\bm{\omega}} > 20.0\text{\si{\radian\per\second}}.
\end{equation}
Upon termination, the reward is replaced by a fixed crash penalty:
\begin{equation}
r_t = -200.
\end{equation}

\subsection{Morphology Feasibility Filtering}
\label{app:feasibility_filtering}
We provide the numerical details of the morphology feasibility filters introduced in~\Cref{sec:method}. For the sampled morphology distribution, motor thrust limits are normalized by vehicle weight. For hexarotors, we set $f_{\min,i}=0$ and $f_{\max,i}=mg/2$ for each motor $i$. For quadrotors, we set $f_{\max,i}=1.5 mg/2$, Using the allocation matrix notation from the main text, motor thrust commands are constrained to lie within
$\bm{f}_{\min} \leq \bm{f} \leq \bm{f}_{\max}$.
First, we test hover feasibility by checking whether there exists a feasible motor command $\bm{u}$ that produces thrust equal to the vehicle weight while generating zero body torque:
\begin{equation}
\|\bm{B}_f \bm{f}\|_2 = mg,
\qquad
\bm{B}_{\tau}\bm{f} = \bm{0},
\qquad
\bm{f}_{\min} \leq \bm{f} \leq \bm{f}_{\max}.
\end{equation}
This feasibility problem is solved numerically with projected gradient-based optimization.

Second, we require a minimum thrust-to-weight margin along the body-frame vertical axis:
\begin{equation}
\frac{f^{\max}_{\mathcal{B},z}}{mg} > b_f,
\end{equation}
where $b_f=2.5$ for hexarotors and $b_f=1.5$ for quadrotors. Finally, we require sufficient angular authority around all body axes. We estimate the maximum achievable angular acceleration from the maximum attainable torque and the system inertia as
\begin{equation}
\bm{\alpha}^{\max} \approx \bm{J}^{-1}\bm{\tau}^{\max}.
\end{equation}
and morphologies are retained only if
\begin{equation}
\alpha^{\max}_x > 16\text{\si{\radian\per\square\second}},\qquad
\alpha^{\max}_y > 16\text{\si{\radian\per\square\second}},\qquad
\alpha^{\max}_z > 16\text{\si{\radian\per\square\second}}.
\end{equation}
After applying all filters, approximately $40$\si{\percent} of sampled hexarotor morphologies and $33$\si{\percent} of sampled quadrotor morphologies are retained.

\subsection{Difficulty-based Morphology Sampling}
\label{app:difficulty_sampling}

During training, we maintain a running estimate of the mean return achieved for each morphology. To increase the sampling frequency of morphologies that are currently harder for the policy, we convert these returns into a difficulty score. For morphology $i$, with running mean return $\bar{R}_i$, we define
\begin{equation}
d_i =
1 -
\frac{
\bar{R}_i - \min_j \bar{R}_j
}{
\max_j \bar{R}_j - \min_j \bar{R}_j
}.
\end{equation}
Thus, morphologies with lower mean return receive higher difficulty scores. At reset, new morphologies are sampled with probability $p_i$ according to
\begin{equation}
p_i =
\frac{\exp(\beta d_i)}
{\sum_j \exp(\beta d_j)}.
\end{equation}
We use $\beta=1$ for hexarotors and $\beta=4$ for quadrotors. This sampling scheme increases the frequency of challenging morphologies while still allowing the policy to train over the full morphology distribution.

\subsection{Architecture}
\label{app:architecture}
\begin{figure}
    \centering
    \includegraphics[width=0.7\linewidth]{ 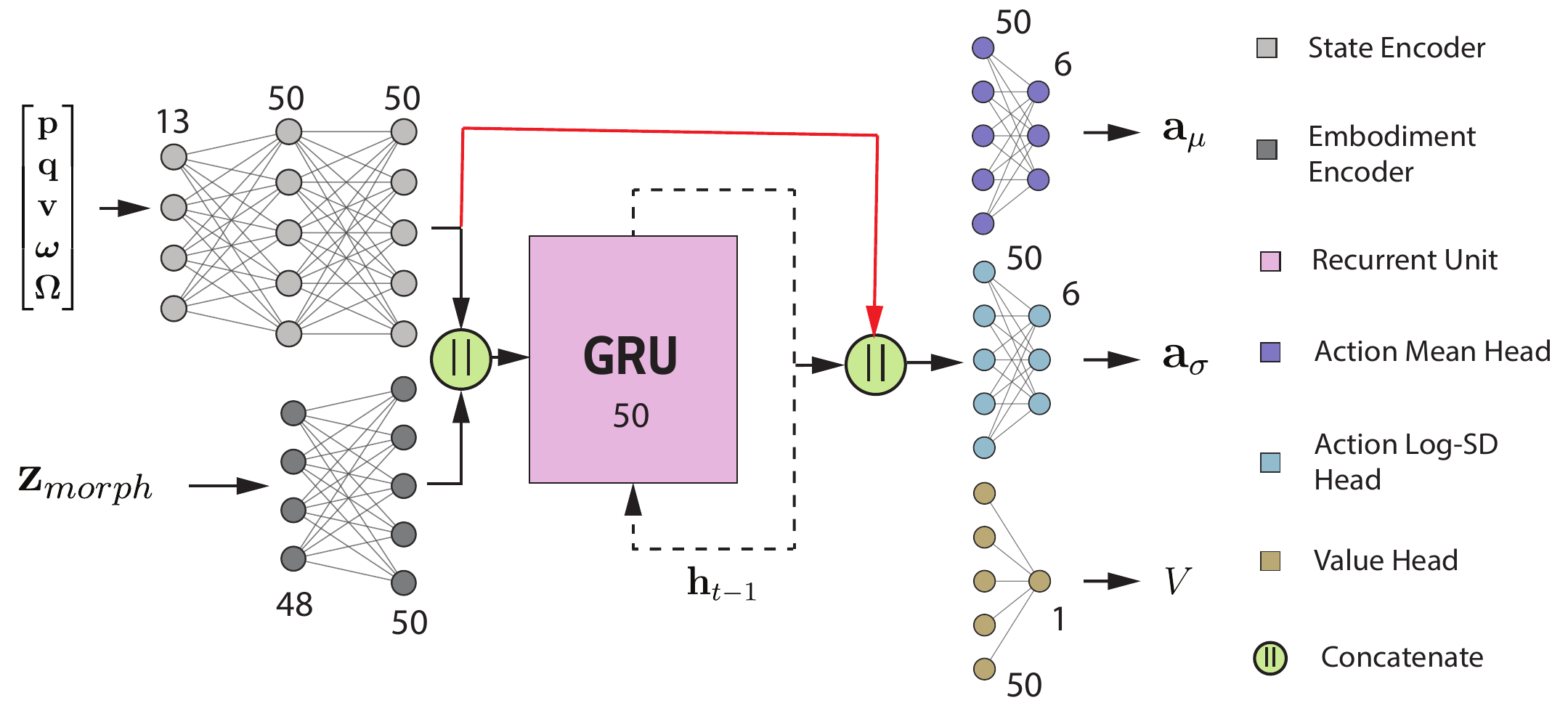}
    \caption{Network architecture.}
    \label{fig:net_architecture}
\end{figure}

The proposed architecture is shown in \Cref{fig:net_architecture}. We use a recurrent policy with a residual feedforward path from the state encoder to the policy head. The recurrent pathway provides memory over past observations, which is useful under observation latency and partial observability induced by unobserved actuator states. The residual pathway preserves direct access to the current state representation, allowing the policy to combine instantaneous feedback with history-dependent features. This design was chosen empirically to improve robustness under noisy observations for sim-to-real transfer.

\section{Quadrotor Results}
\label{app:quadrotors}

We repeat the simulation study on a population of quadrotor morphologies to verify that the proposed descriptor remains effective beyond the hexarotor setting considered in the main experiments. The quadrotor population is sampled using the same procedure as in the main text, but now each motor is sampled within a $\zeta = 360\text{\si{\degree}}/4 = 90\text{\si{\degree}}$ sector. We increase the maximum thrust per motor by a factor of $6/4$ relative to the hexarotor experiments, keeping the total available thrust comparable across the two morphology families. All other training and evaluation settings are kept unchanged. The evaluation protocol is identical: policies are evaluated on Lissajous trajectory tracking over sampled morphologies, and we report success rate together with position and velocity RMSE.

\begin{figure}[htbp]
    \centering

    \begin{subfigure}{0.535\linewidth}
        \centering
        \includegraphics[width=\linewidth]{ 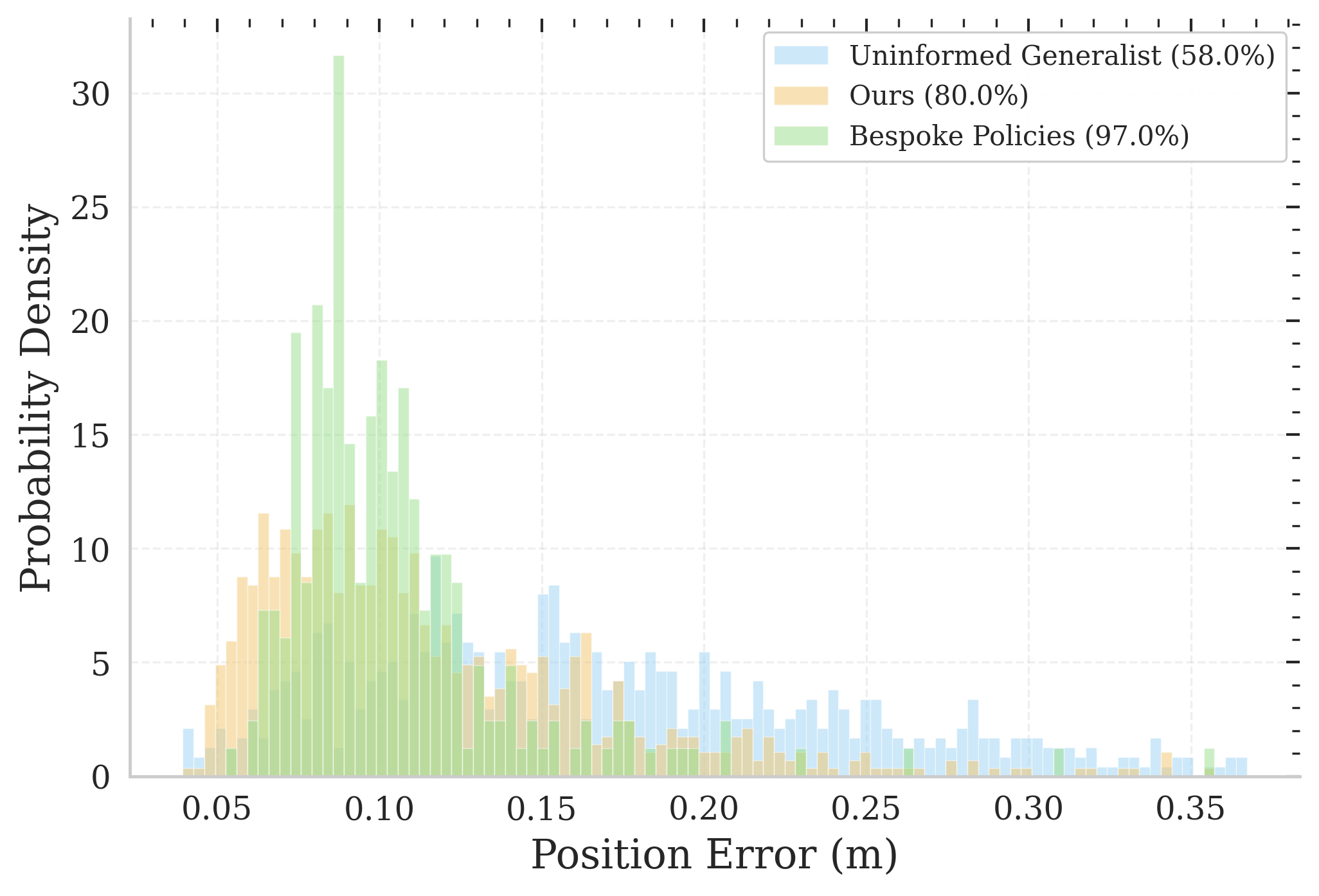}
        \caption{}
        \label{fig:rmse_quads}
    \end{subfigure}
    \hfill
    \begin{subfigure}{0.45\linewidth}
        \centering
        \includegraphics[width=\linewidth]{ 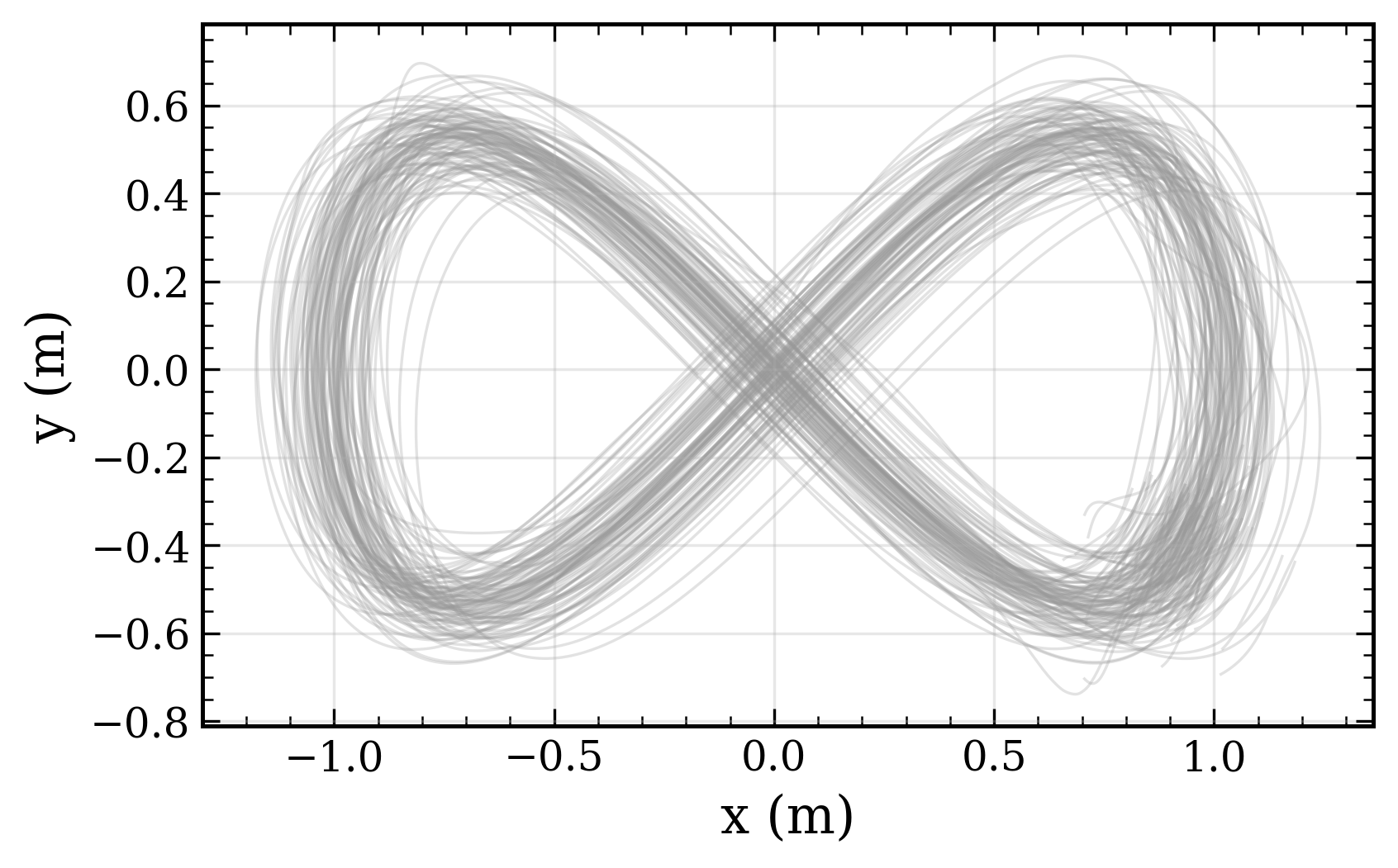}
        \caption{}
        \label{fig:traj_quads}
    \end{subfigure}

    \caption{Simulation results on quadrotor morphologies. (a) Distribution of position tracking errors during the Lissajous trajectory task for specialist policies, the uninformed generalist, and the proposed embodiment-informed generalist. (b) Trajectories generated by the embodiment-informed generalist across 100 randomly sampled quadrotor embodiments.}
    \label{fig:lissajous_results_quads}
\end{figure}

The quantitative results are reported in \Cref{tab:sim_results_compare_quads} and \Cref{fig:lissajous_results_quads}. The embodiment-conditioned policy substantially outperforms the uninformed generalist, improving the success rate from $58$\si{\percent} to $80$\si{\percent} and reducing both position and velocity RMSE, while the bespoke policies still provide an upper bound, reaching $97$\si{\percent} success rate. In \Cref{fig:lissajous_results_quads}, we also demonstrate simulated trajectories, using our policy. The quadrotor results provide an opportunity to qualitatively compare our approach to RAPTOR~\cite{raptor} which otherwise is limited to planar, symmetric X-quadrotor platforms. Accordingly, we evaluate our policy on a distribution of planar, symmetric X-quadrotors generated following the geometry parameterization of RAPTOR~\cite{raptor}, while clamping mass and actuator dynamics to the range covered by our training distribution. The controller achieves a 96\% success rate, with position RMSE of $0.065 \pm 0.023$~m and velocity RMSE of $0.046 \pm 0.038$~m/s. These results highlight the increased performance of our generalist on conventional planar systems, while our policy further supports a wider set of systems including non-planar and asymmetric designs.

%The lower success rate observed for quadrotors motivated additional investigation of the embodiment descriptor. We found that replacing the normalized allocation matrix, $\bm{B}_{\mathrm{norm}}$, with its pseudo-inverse, $\bm{B}_{\mathrm{norm}}^{\dagger}$, improves the success rate from $80$\si{\percent} to $89$\si{\percent}. We hypothesize that, unlike in the hexarotor setting, $\bm{B}_{\mathrm{norm}}^{\dagger}$ may provide a numerically more convenient representation for policy learning in quadrotors.
%, as it directly relates desired generalized forces to motor commands.

\begin{table}[t]
\centering
\caption{Performance comparison of control policies on the quadrotor morphology distribution. Success rate (SR), position RMSE, and velocity RMSE are reported over the Lissajous trajectory tracking task.}
\label{tab:sim_results_compare_quads}
\small
\setlength{\tabcolsep}{3pt}
\begin{tabular}{lc|cc}
\toprule
\textbf{Method} &
\textbf{SR (\si{\percent})} &
\textbf{$\bm{p}$ RMSE (\si{\meter})} &
\textbf{$\bm{v}$ RMSE (\si{\meter\per\second})} \\
\midrule 
Bespoke Policies
& 97 & $0.086\pm 0.035$ & $0.068 \pm 0.077$\\
Uninformed Generalist
& 58 & $0.181 \pm 0.080$ & $0.066 \pm 0.092$ \\
Ours
& 80 & $0.105 \pm 0.055$ & $0.039 \pm 0.014$ \\
% Ours w $\bm{B}_{\mathrm{norm}}^{\dagger}$
% & 89 & $0.108 \pm 0.063$ & $0.051 \pm 0.060$ \\

\bottomrule
\end{tabular}
\end{table}

%We also evaluate this policy on a distribution of planar, symmetric quadrotors as described in RAPTOR and we have comparable performance. 

% \begin{table}[t]
% \centering
% \caption{Comparison with raptor on their sistribution}
% \label{tab:sim_results_compare_quads_raptor}
% \small
% \setlength{\tabcolsep}{3pt}
% \begin{tabular}{lc|cc}
% \toprule
% \textbf{Method} &
% \textbf{Success Rate (\%)} &
% \textbf{$\bm{p}$ RMSE (m)} &
% \textbf{$\bm{v}$ RMSE (m/s)} \\
% \midrule 
% Ours
% & 100 & $0.038\pm 0.001$ & $0.033 \pm 0.001$\\
% Raptor
% &  & & \\

% \bottomrule
% \end{tabular}
% \end{table}

\section{The Three Representative Embodiments}
\label{app:representatives}
\begin{figure}[t]
    \centering
    \begin{subfigure}[t]{0.4\linewidth}
        \centering
        \includegraphics[width=\linewidth]{ 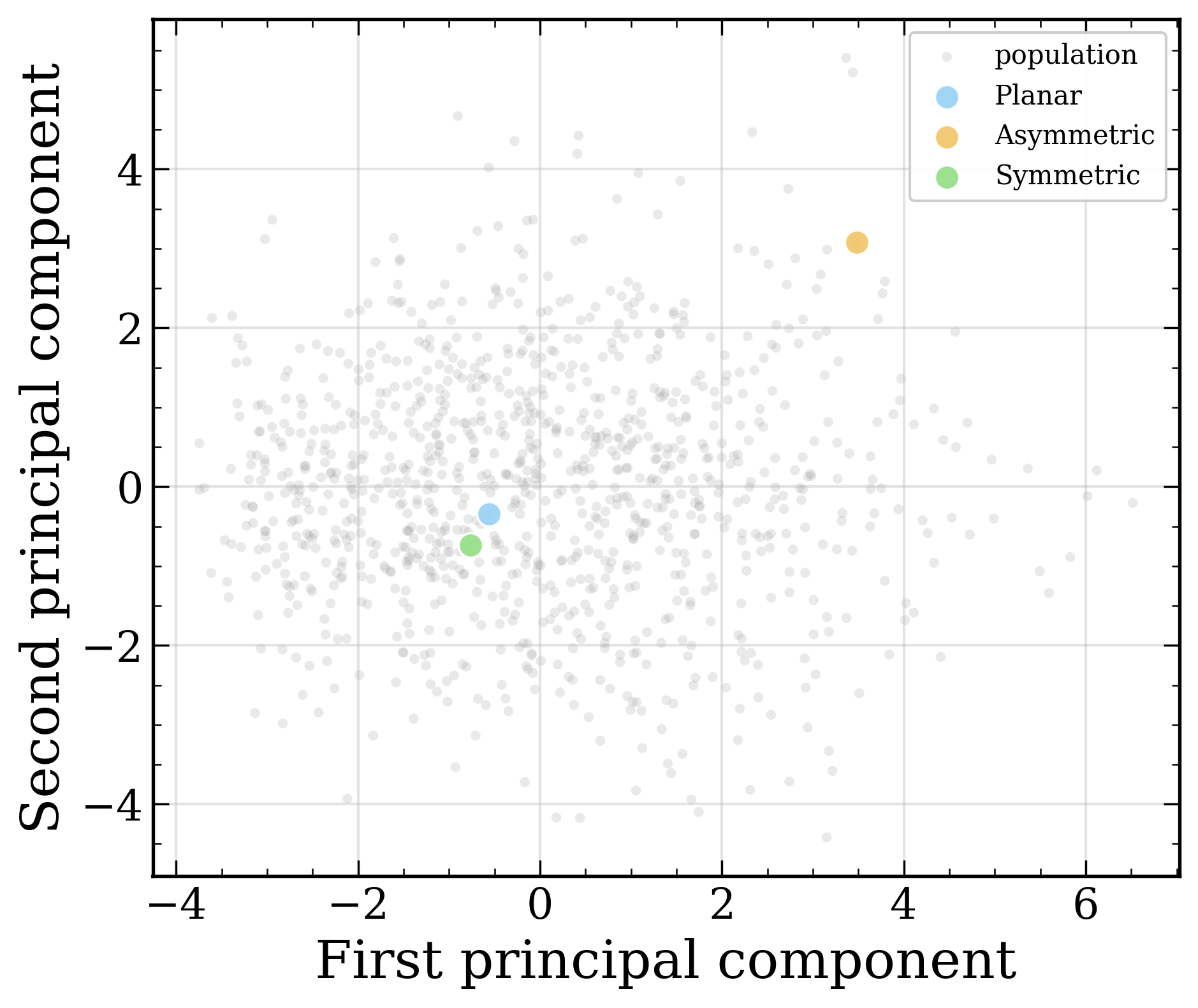}
        \caption{Morphology descriptor}
        \label{fig:pca_morphology}
    \end{subfigure}
    \hfill
    \begin{subfigure}[t]{0.4\linewidth}
        \centering
        \includegraphics[width=\linewidth]{ 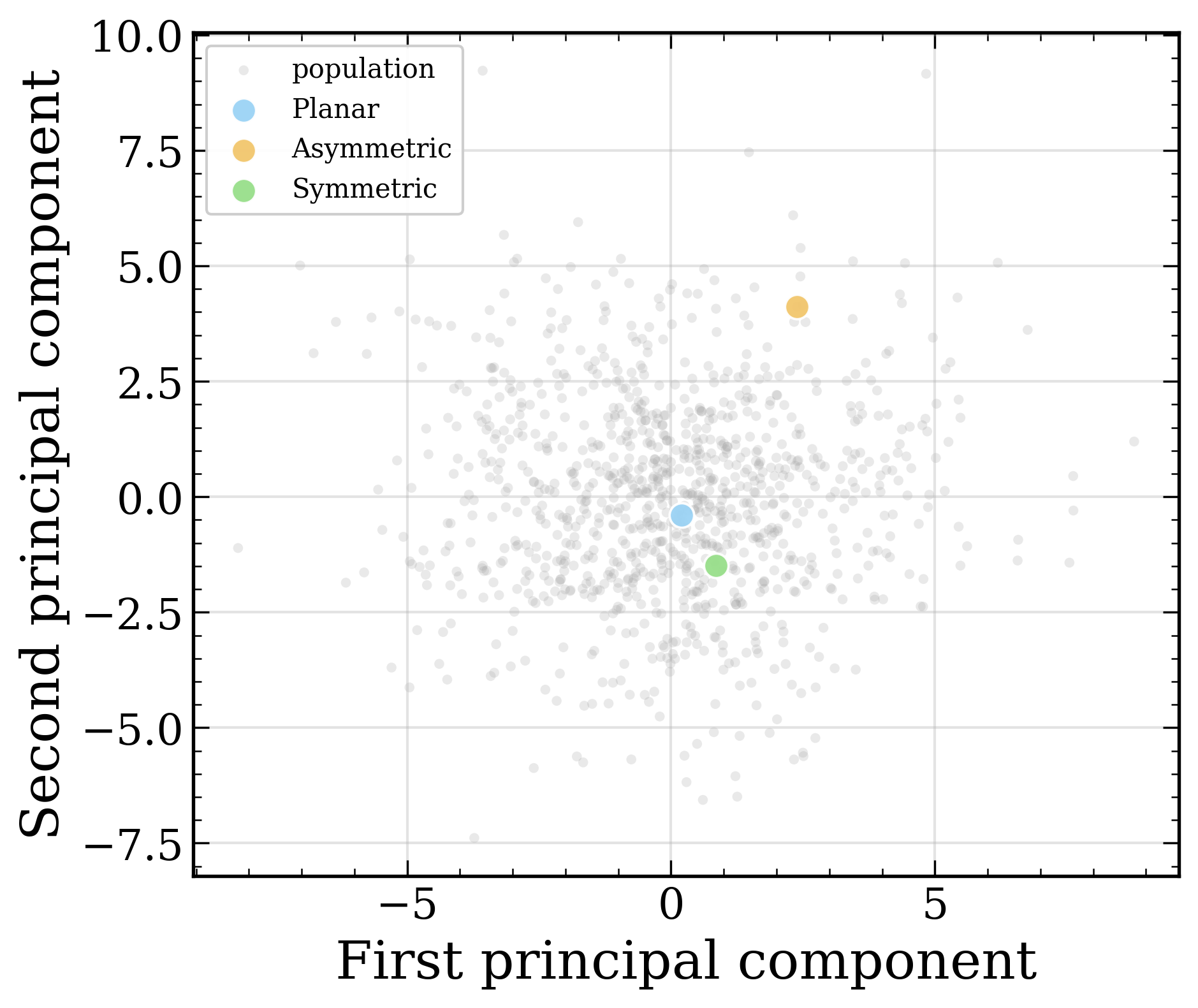}
        \caption{Policy embedding}
        \label{fig:pca_policy_embedding}
    \end{subfigure}
    \caption{Two-dimensional projections of morphology-related representations.}
    \label{fig:pca_representations}
\end{figure}

 \Cref{tab:3emb_params} reports the physical parameters of the three representative hexarotor embodiments used for real-world and qualitative evaluations: the planar, the non-planar symmetric and the asymmetric non-planar morphology. These platforms were selected to cover qualitatively different regions of the morphology distribution, ranging from a conventional planar design to configurations with tilted motors and non-planar motor placements. \Cref{fig:pca_representations} visualizes where these three morphologies lie relative to the sampled population. On the left, we show a PCA projection of our morphology descriptor, while on the right a PCA projection of the learned policy embedding.  

\begin{table}[]
    \centering
    \caption{Detailed embodiment parameters of the $3$ systems used for real world evaluation.}
    \scriptsize
    \begin{tabular}{l|l|l|l}
            \toprule
             & \textbf{Planar} & \textbf{Symmetric} & \textbf{Asymmetric} \\
             \midrule
             \multicolumn{4}{c}{\textbf{Mass (\si{\kilogram})}} \\
             \midrule
            $m$ & $0.421$ & $0.411$  & $ 0.402$  \\
            \midrule
            \multicolumn{4}{c}{\textbf{Motor Positions (\si{\meter}) (with respect to $\mathcal{B}$)}} \\
            \midrule
            motor $1$ & $[ 0.116, -0.072,  0.000]$ & $[ 0.106, -0.061,  0.04    ]$  & $[ 0.0967, -0.088, -0.049]$  \\
            motor $2$ & $[ 0.000, -0.137,  0.000]$  & $[ 0.      , -0.123, -0.03    ]$ & $[ 0.016, -0.132,  0.042]$  \\
            motor $3$ & $[-0.116, -0.072,  0.000]$  & $[-0.106, -0.061,  0.04    ]$ & $[-0.093, -0.065 , -0.057]$  \\
            motor $4$ & $[-0.116,  0.072,  0.000]$  & $[-0.106, 0.061,  0.04    ]$ & $[-0.108,  0.061,  0.063]$  \\
            motor $5$ & $[0.000,  0.137,  0.000]$  & $[ 0.      ,  0.122, -0.03    ]$ & $[-0.011,  0.138,  0.0133]$  \\
            motor $6$ & $[0.116,  0.072,  0.000]$  & $[ 0.106,  0.061,  0.04    ]$ & $[ 0.127,  0.037,  0.022]$  \\
            \midrule
            \multicolumn{4}{c}{\textbf{Motor Orientations (\si{\degree}) (xyz-euler angles)}}\\
            \midrule
            $\phi_1$ & $[0,0,0]$ & $[ 14.4, -25.3,  0.0]$  & $[1.05, -13.3, 0.0]$ \\
            $\phi_2$ & $[0,0,0]$ & $[-13.5,  25.7,  0.0]$  & $[-10.0, -19.5, 0.0]$ \\
            $\phi_3$ & $[0,0,0]$ & $[-28.9,  -0.5,  0.0]$  & $[4.9, -1.4, 0.0]$ \\
            $\phi_4$ & $[0,0,0]$ & $[ 28.9,  -0.5,  0.0]$  & $[5.8, -7.9, 0.0]$ \\
            $\phi_5$ & $[0,0,0]$ & $[ 13.5,  25.7,  0.0]$  & $[-11.2, -0.2, 0.0]$ \\
            $\phi_6$ & $[0,0,0]$ & $[-14.4, -25.3,  0.0]$  & $[4.2, -3.0, 0.0]$ \\
            \midrule
            \multicolumn{4}{c}{\textbf{Thrust Coefficient (\si{\newton\square\second})}} \\
            \midrule
            propeller $1...6$  & $1.286 \times 10^{-5}$ & $2.308 \times 10^{-5}$ & $1.286 \times 10^{-5}$ \\
            \midrule
            \multicolumn{4}{c}{\textbf{Time Constant (\si{\second})}} \\
            actuator $1...6$ & $0.047$ & $0.05$ & $0.047$ \\
            \midrule
            \multicolumn{4}{c}{\textbf{Torque-to-Thrust Ratio (\si{\meter})}}\\
            \midrule
            propeller $1...6$ & $0.01$ & $0.015$ & $0.01$ \\
            \bottomrule
    \end{tabular}
    \normalsize
    \label{tab:3emb_params}
\end{table}

\normalsize

\section{Analysis of Failing Embodiments}
\label{app:failing_embodiments}

\begin{figure}[t]
    \centering
    \begin{subfigure}[t]{0.32\linewidth}
        \centering
        \includegraphics[width=\linewidth]{ 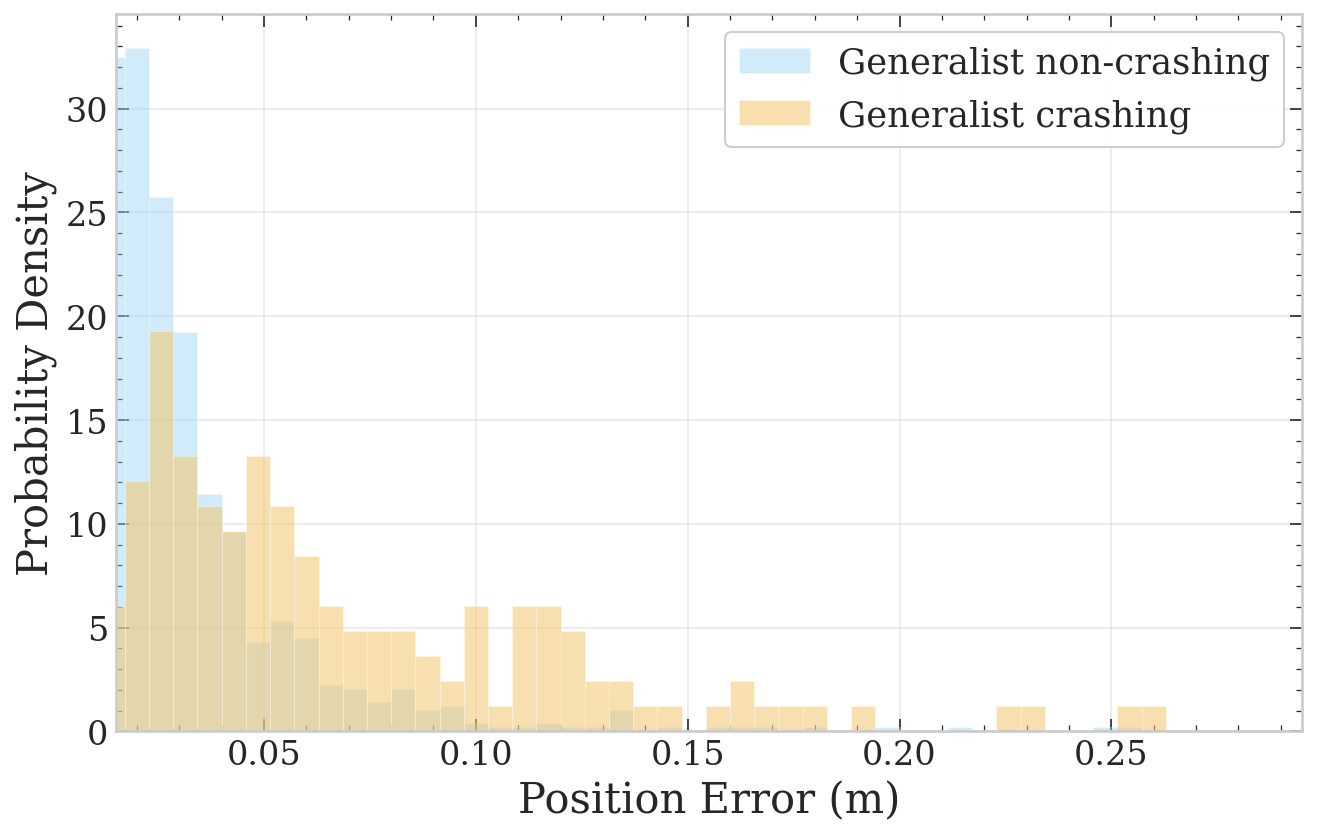}
        \caption{Position error}
        \label{fig:pos_fe_cor}
    \end{subfigure}
    \hfill
    \begin{subfigure}[t]{0.32\linewidth}
        \centering
        \includegraphics[width=\linewidth]{ 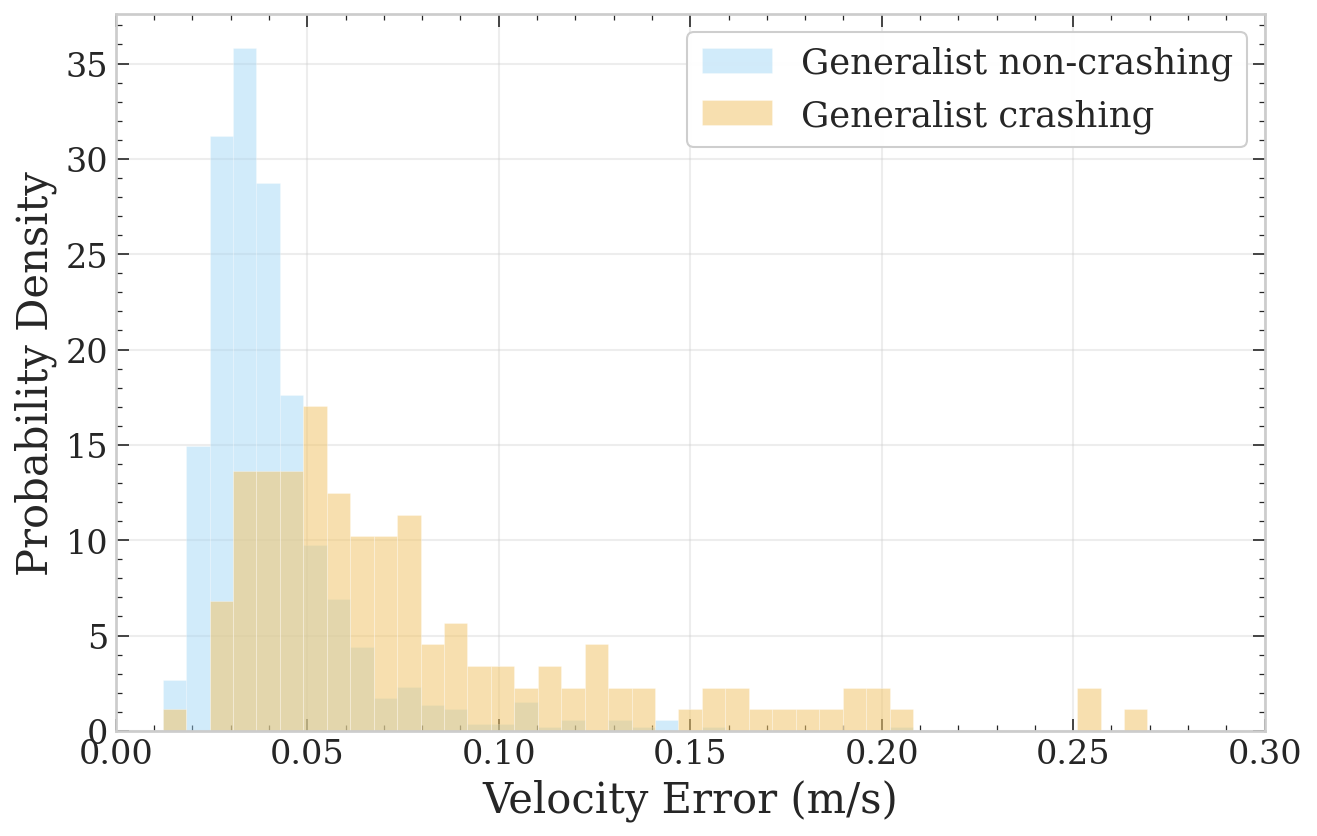}
        \caption{Velocity error}
        \label{fig:vel_fe_cor}
    \end{subfigure}
    \hfill
    \begin{subfigure}[t]{0.32\linewidth}
        \centering
        \includegraphics[width=\linewidth]{ 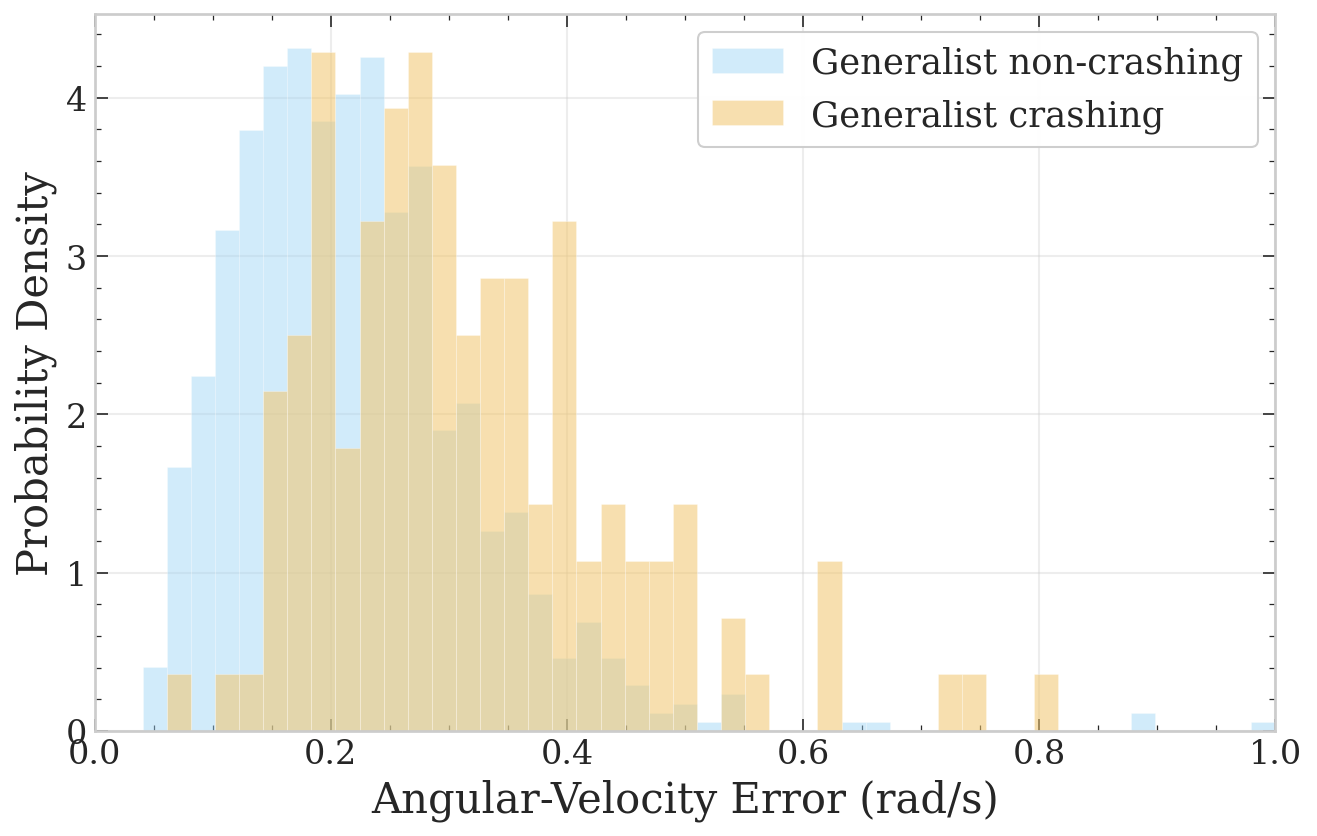}
        \caption{Angular-velocity error}
        \label{fig:angvel_fe_cor}
    \end{subfigure}

\caption{Tracking-error distributions of bespoke policies for embodiments where the embodiment-informed generalist succeeds and fails. Generalist failures are associated with higher position, velocity, and angular-velocity errors under bespoke control}
    \label{fig:failing_embodiments}
\end{figure}

The morphology distribution considered in this work is intentionally broad and includes highly non-standard multirotor configurations. As a result, the zero-shot success rate of a single generalist policy is not perfect, reaching $85\%$ on hexarotors and $80\%$ on quadrotors. However, these failing embodiments are not uncontrollable, since bespoke policies trained on the same configurations can control them. To better understand these failures, we compare the generalist success labels with the tracking errors achieved by bespoke policies on the same embodiments. As shown in \Cref{fig:failing_embodiments}, embodiments on which the generalist fails tend to have higher bespoke position, velocity and angular-velocity errors. This suggests that generalist failures are correlated with the dynamic characteristics of the morphology, rather than being uniformly distributed across the test set.

We further evaluate the generalist on a restricted subset of morphologies with non-tilted motors in \Cref{tab:sim_results_compare_planar}. This subset follows the same morphology family considered in the main experiments, but fixes $\delta = 45^\circ$ and $\phi = 0^\circ$, such that all motors are upright and lie in the $x$--$y$ plane. On this subset, the generalist achieves $100\%$ success rate and tracking performance comparable to bespoke policies. This indicates that the proposed controller is reliable on standard multirotor designs, while failures mainly arise in the broader morphology distribution containing more challenging non-planar or strongly asymmetric configurations.

\begin{table}[t]
\centering
\caption{Performance comparison between the proposed embodiment-informed generalist and bespoke policies on the planar morphology subset. Success rate (SR), position RMSE, and velocity RMSE are reported for the Lissajous trajectory tracking task.}
\vspace{3mm}
\label{tab:sim_results_compare_planar}
\small
\setlength{\tabcolsep}{3pt}
\begin{tabular}{lc|cc}
\toprule
\textbf{Method} &
\textbf{SR (\si{\percent})} &
\textbf{$\bm{p}$ RMSE (\si{\meter})} &
\textbf{$\bm{v}$ RMSE (\si{\meter\per\second})} \\
\midrule 
Ours
& 100 & $0.046\pm 0.011$ & $0.038 \pm 0.010$\\
Bespoke
& 100 & $0.019\pm 0.024$ & $0.003 \pm 0.036$ \\

\bottomrule
\end{tabular}
\end{table}

\section{Sensitivity Analysis}
\begin{figure*}[t]
    \centering
    \setlength{\tabcolsep}{1pt}
    \setlength{\abovecaptionskip}{2pt}
    \setlength{\belowcaptionskip}{0pt}
    \begin{subfigure}[t]{0.245\textwidth}
        \centering
        \includegraphics[width=\linewidth]{ 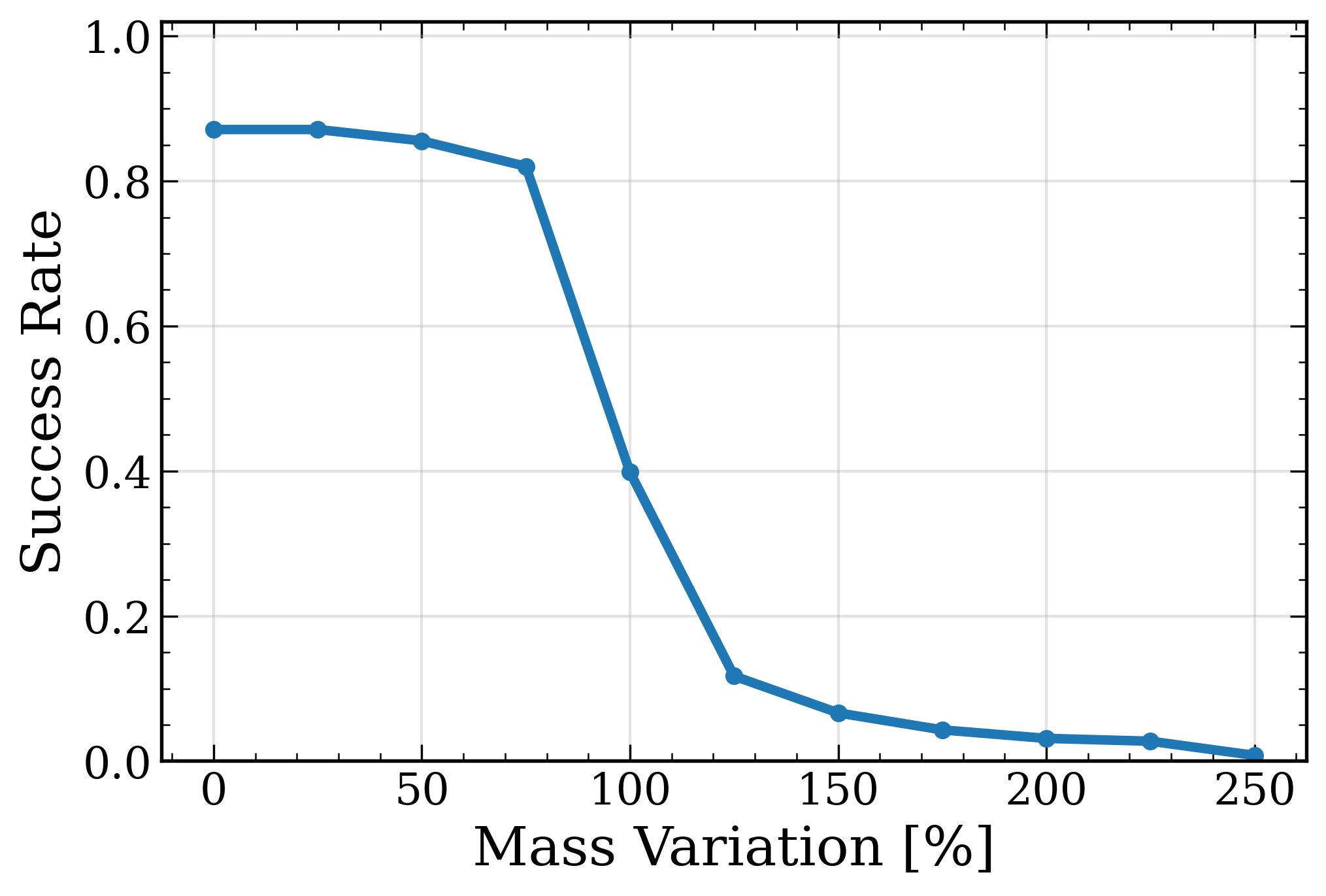}
        \label{fig:success_mass}
    \end{subfigure}
    \hfill
    \begin{subfigure}[t]{0.245\textwidth}
        \centering
        \includegraphics[width=\linewidth]{ 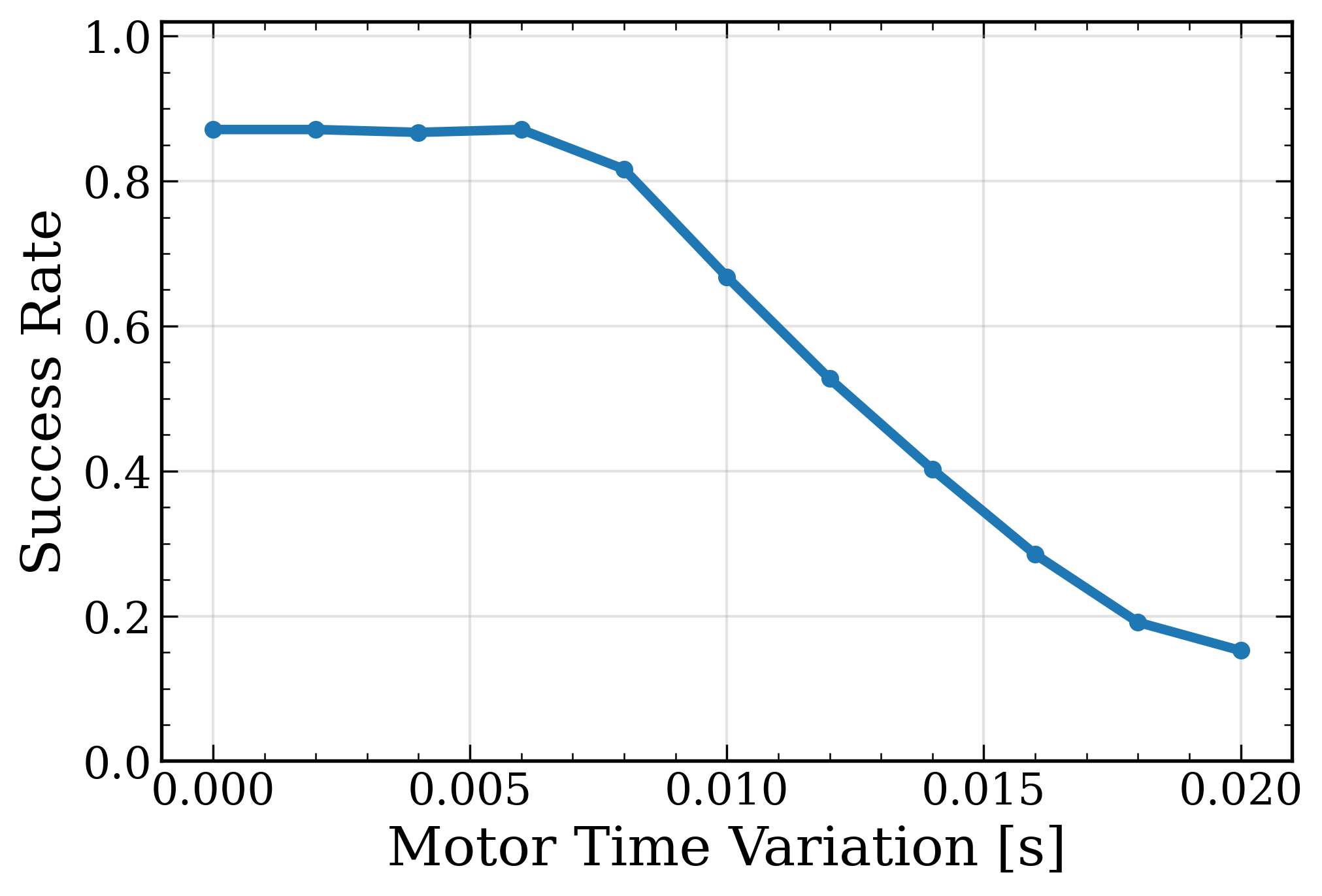}
        \label{fig:success_motor_time}
    \end{subfigure}
    \hfill
    \begin{subfigure}[t]{0.245\textwidth}
        \centering
        \includegraphics[width=\linewidth]{ 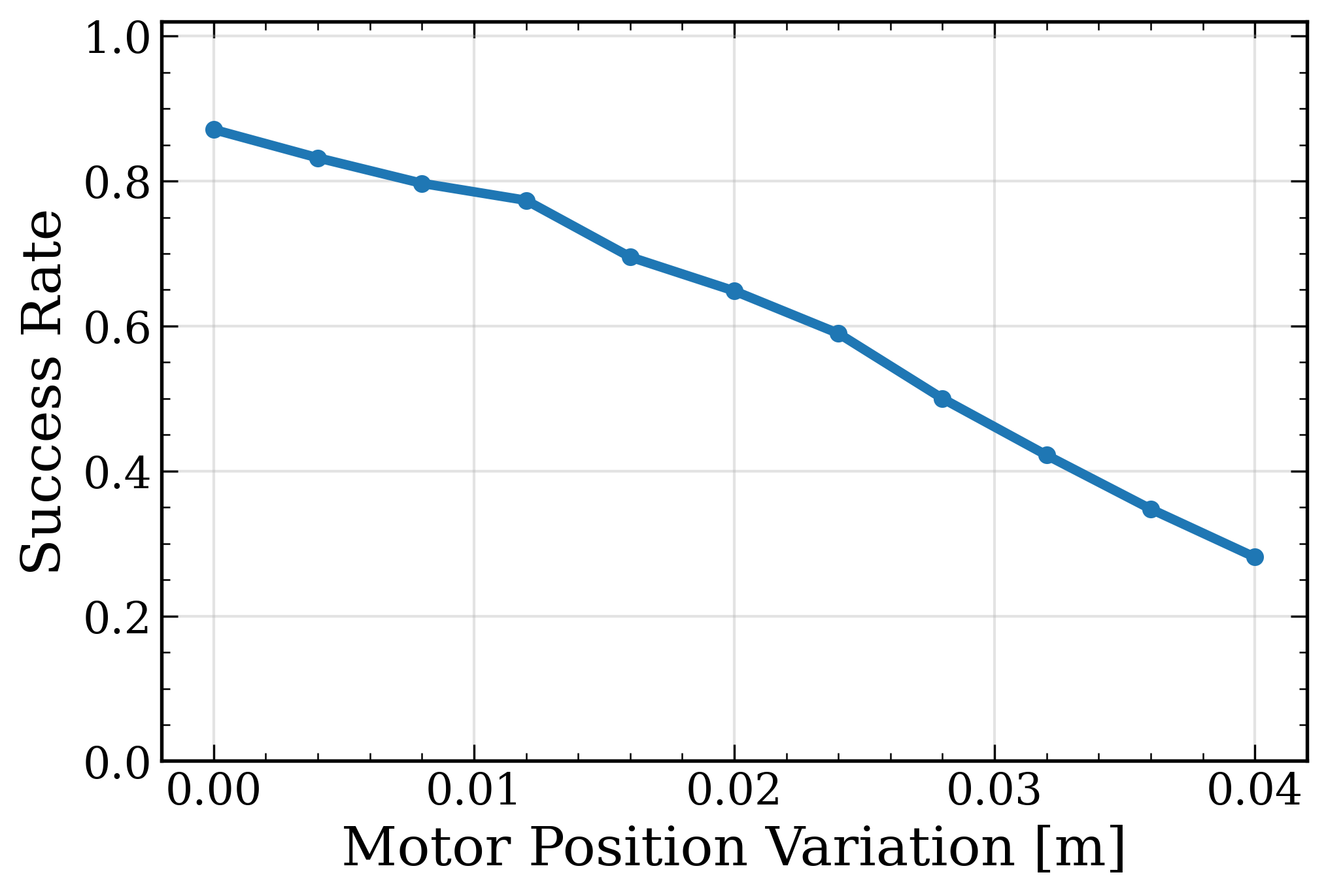}
        \label{fig:success_motor_pos}
    \end{subfigure}
    \hfill
    \begin{subfigure}[t]{0.245\textwidth}
        \centering
        \includegraphics[width=\linewidth]{ 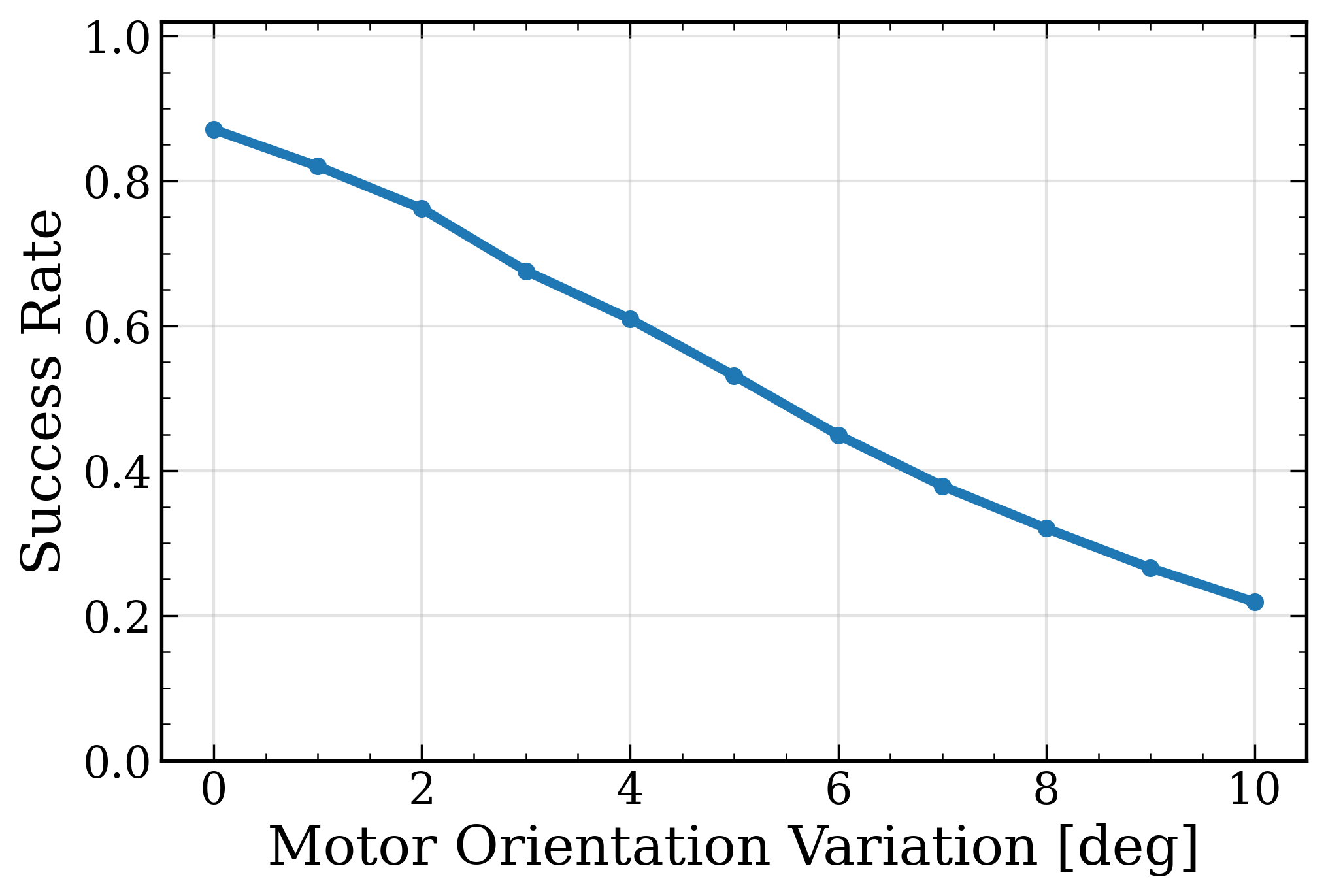}
        \label{fig:success_motor_ori}
    \end{subfigure}

    \caption{Success rate evolution with increasing noise on the embodiment information provided to the generalist. Mass is perturbed proportionally relative to its nominal value. Motor time constants are perturbed uniformly within the range shown on the x-axis in seconds, motor positions in meters and motor orientations in degrees.}
    \label{fig:emb_obs_noise}
\end{figure*}

We evaluate the robustness of the generalist policy to errors in the embodiment descriptor. At evaluation time, we perturb one class of the embodiment parameters without informing the policy, measuring sensitivity to a mismatch between the assumed and true morphology. Perturbations are scaled relative to nominal parameter ranges: mass is perturbed proportionally to its nominal values, while motor time constants, motor positions and orientations are perturbed additively in seconds, meters and degrees, respectively. As shown in \Cref{fig:emb_obs_noise}, the policy is highly robust to mass perturbations and motor time-constant perturbations up to moderate noise levels, after which performance drops. The policy is more sensitive to geometric perturbations as expected, since they directly affect the achievable force--torque directions.

\begin{comment}

\begin{figure}
    \centering
    \includegraphics[width=0.7\linewidth]{ pics/success_rate_comparison_all_components.pdf}
    \caption{Success rate evolution with increasing noise on the embodiment information provided to the generalist. A noise scale of $s$ corresponds to mass perturbations of $0.05s$ relative to nominal mass, motor-position perturbations of $\pm 0.002s\,\mathrm{m}$, motor-orientation perturbations of $\pm s^\circ$, and motor time-constant perturbations of $0.1s$ relative to nominal time constants.}
    \label{fig:emb_obs_noise}
\end{figure}
    
\end{comment}

\section{Step Responses under Noisy Sim-to-Real Conditions}
\label{app:noisy_step_response}

\begin{figure*}[t]
    \centering
    \setlength{\abovecaptionskip}{2pt}
    \setlength{\belowcaptionskip}{0pt}
    \begin{subfigure}[t]{0.95\textwidth}
        \centering
        \includegraphics[width=\linewidth]{ 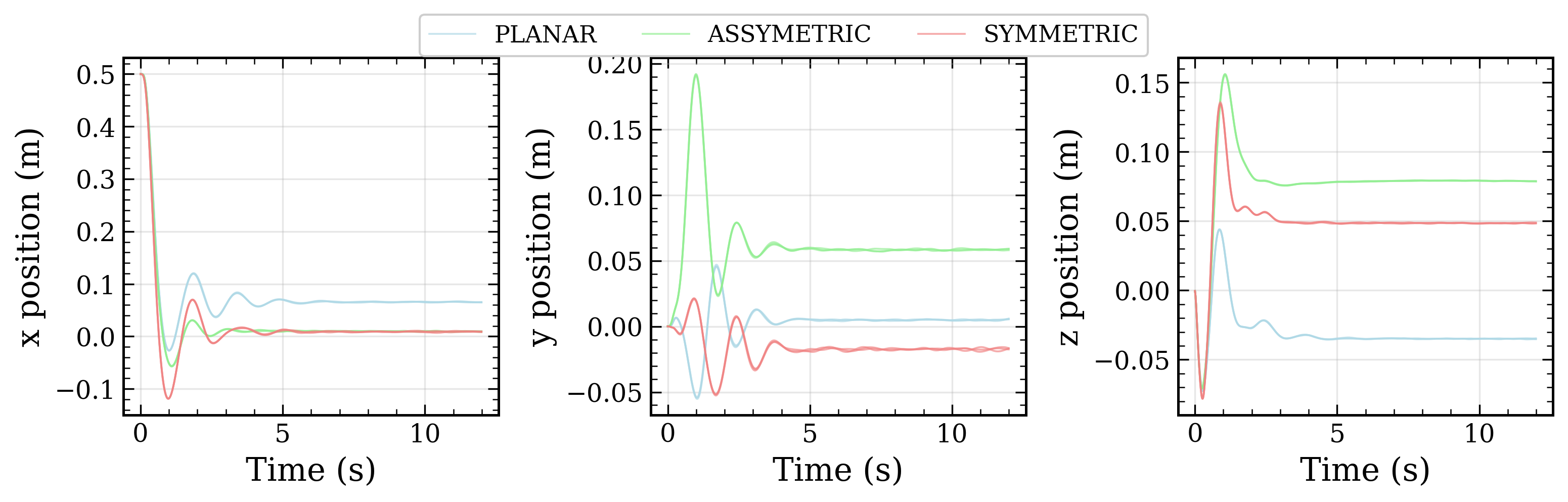}
        \caption{Position}
        \label{fig:state_position}
    \end{subfigure}
    \vspace{0.3em}
    \begin{subfigure}[t]{0.95\textwidth}
        \centering
        \includegraphics[width=\linewidth]{ 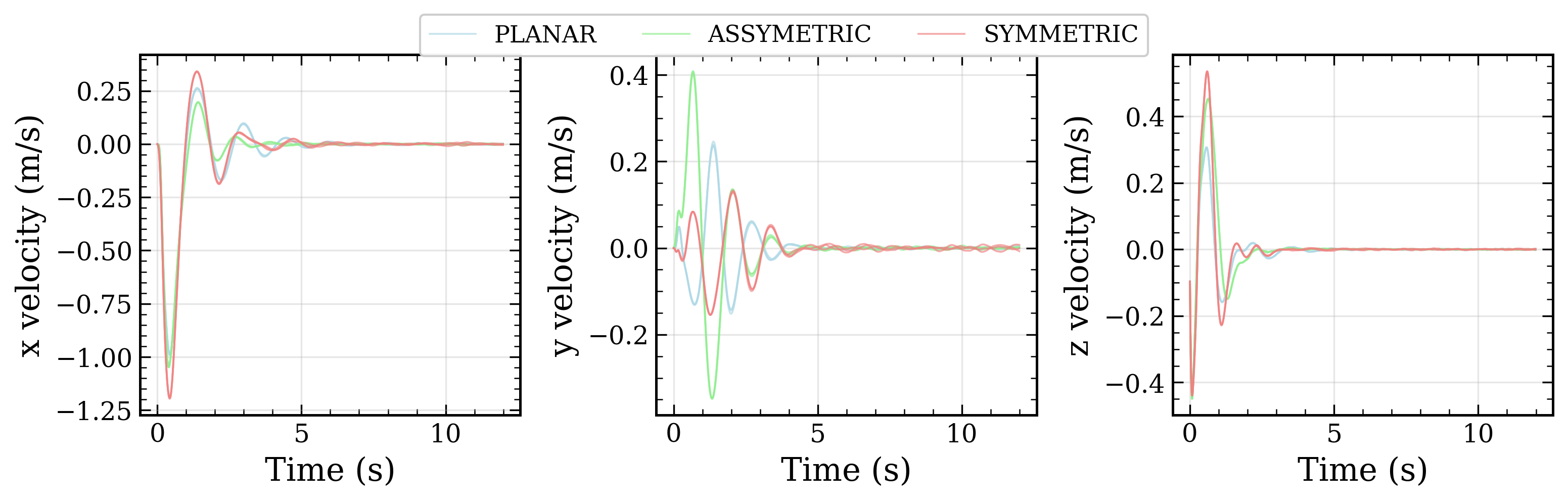}
        \caption{Velocity}
        \label{fig:state_velocity}
    \end{subfigure}
    \caption{Position and velocity step responses of the policy deployed under sim-to-real noise and latency conditions. The three representative embodiments are excited with a $0.5\,\mathrm{m}$ step position command along the $x$ axis. Three independent rollouts per embodiment are shown.}
    \label{fig:rollout_traces_pos_vel}
\end{figure*}

\begin{figure*}[t]
    \centering
    \setlength{\abovecaptionskip}{2pt}
    \setlength{\belowcaptionskip}{0pt}
    \begin{subfigure}[t]{0.95\textwidth}
        \centering
        \includegraphics[width=\linewidth]{ 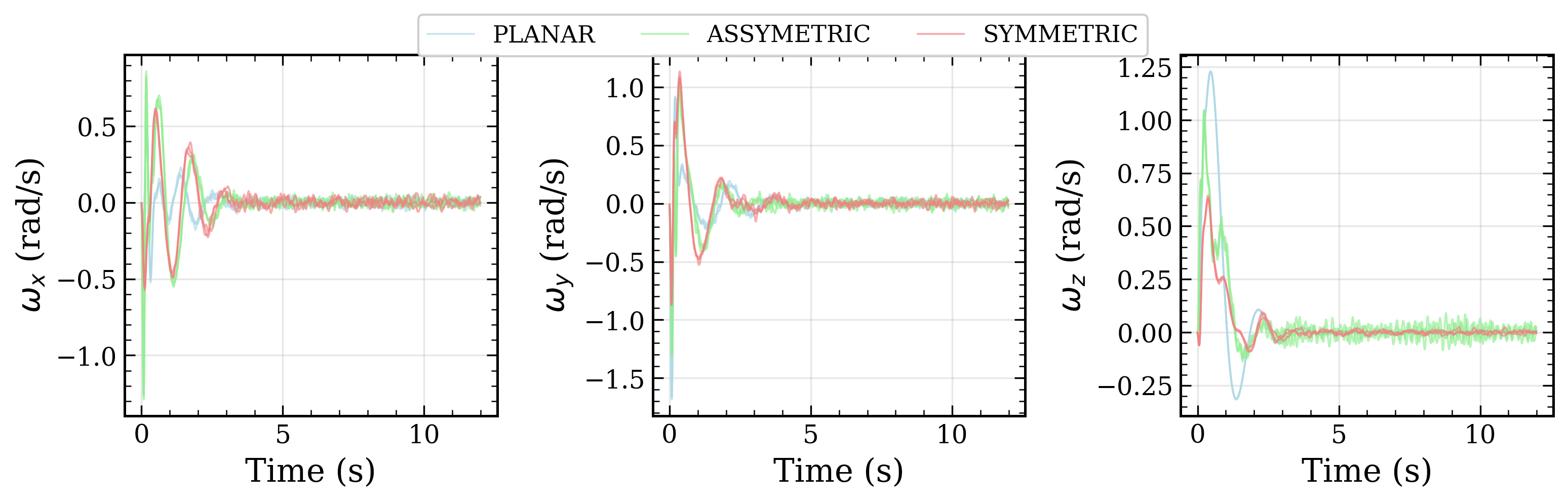}
        \caption{Angular velocity}
        \label{fig:state_angvel}
    \end{subfigure}
    \vspace{0.3em}
    \begin{subfigure}[t]{0.95\textwidth}
        \centering
        \includegraphics[width=\linewidth]{ 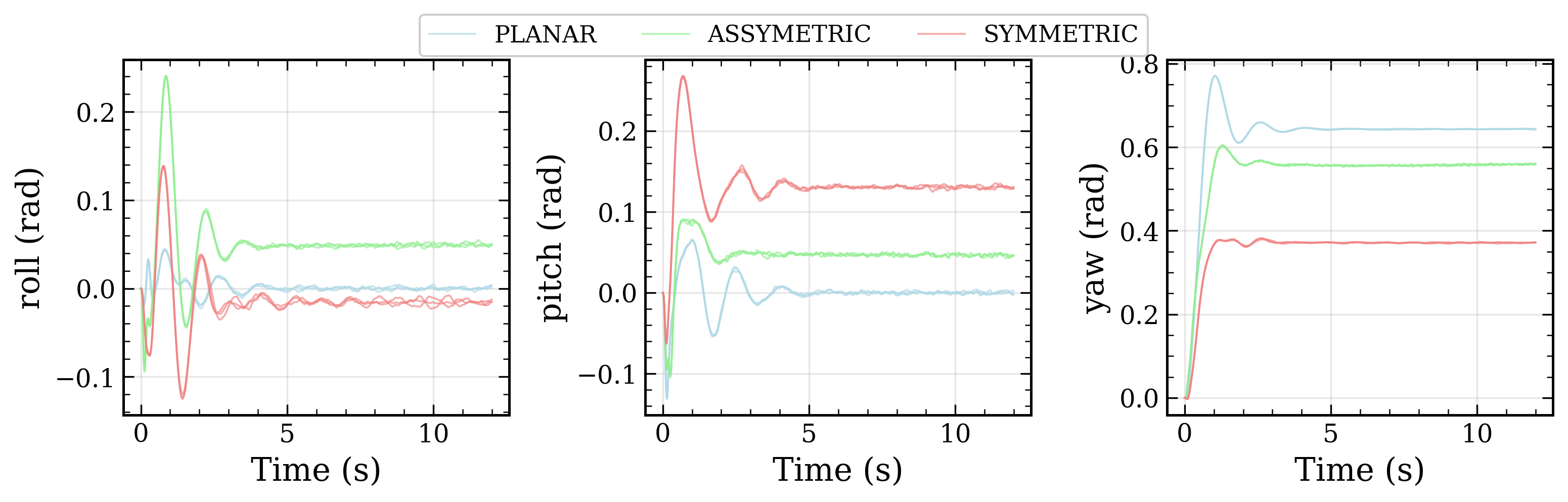}
        \caption{Euler angles}
        \label{fig:state_angles}
    \end{subfigure}
    \caption{Angular velocity and Euler angle step responses of the policy deployed under sim-to-real noise and latency conditions.}
    \label{fig:rollout_traces_ang}
\end{figure*}

\begin{figure*}[t]
    \centering
    \setlength{\abovecaptionskip}{2pt}
    \setlength{\belowcaptionskip}{0pt}
    \begin{subfigure}[t]{0.95\textwidth}
        \centering
        \includegraphics[width=\linewidth]{ 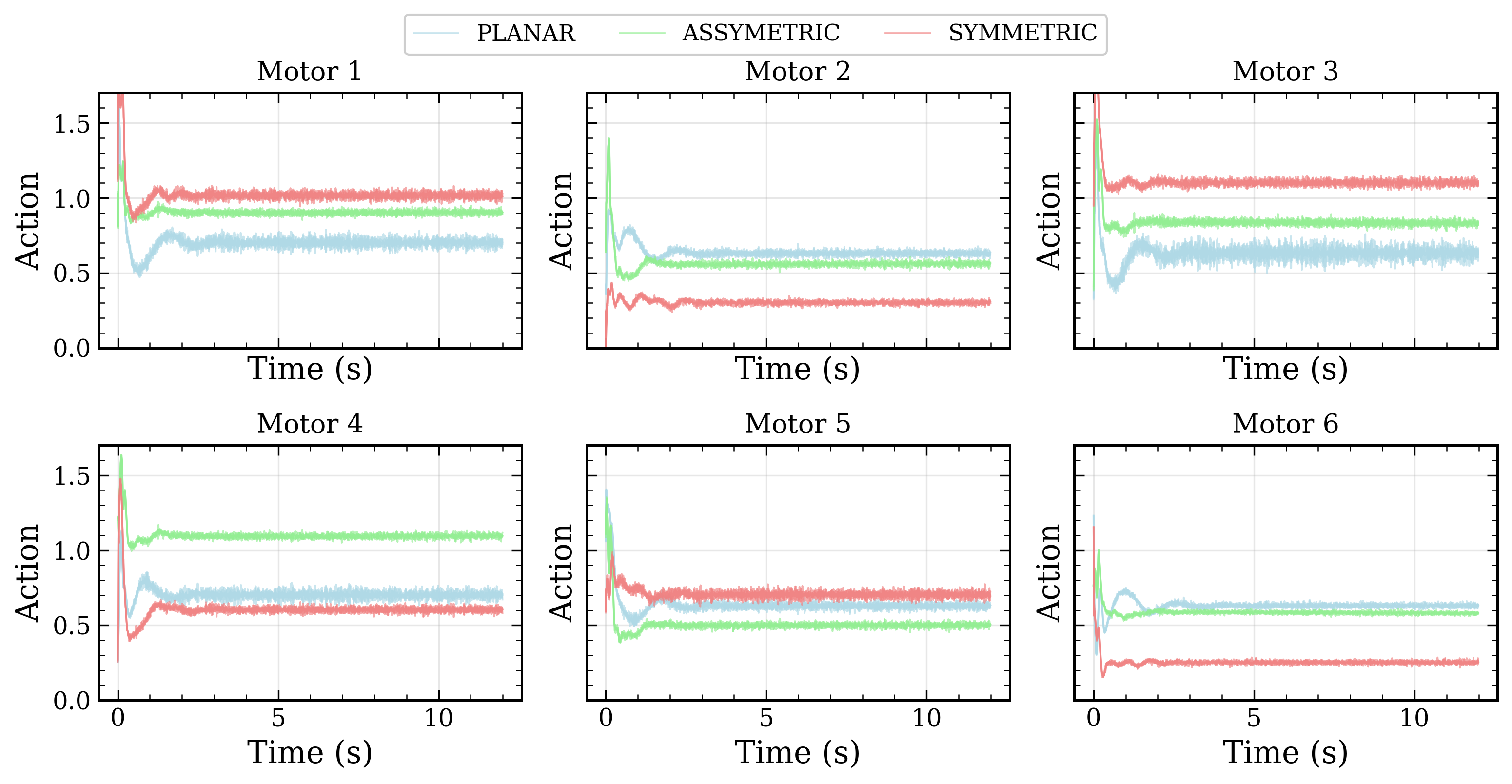}
        \caption{Actions}
        \label{fig:state_actions}
    \end{subfigure}
    \caption{Motor thrust commands (actions) of the policy deployed under sim-to-real noise and latency conditions.}
    \label{fig:rollout_traces_actions}
\end{figure*}

We further evaluate the behavior of the generalist controller used for deployment on the real platforms. Specifically, we test the policy in simulation under the same observation-noise and latency setting used for sim-to-real training. The three representative systems are excited with a $0.5\,\mathrm{m}$ step command along the $x$-axis, and we report the resulting position in \Cref{fig:state_position}, velocity in  \Cref{fig:state_velocity}, angular velocity in \Cref{fig:state_angvel}, Euler angles in \Cref{fig:state_angles}, and policy actions in \Cref{fig:rollout_traces_actions}. For each embodiment, we show three independent rollouts with different realizations of the observation noise.

\end{document}